\documentclass[10pt,twocolumn,letterpaper]{article}

\usepackage{iccv}
\usepackage{times}
\usepackage{epsfig}
\usepackage{graphicx}
\usepackage{amsmath}
\usepackage{amssymb}
\usepackage{bbm}
\usepackage{booktabs}
\usepackage{bm}
\usepackage{multirow}
\usepackage{microtype}
\usepackage[T1]{fontenc}
\usepackage{soul}

\usepackage{xcolor}
\usepackage{listings}
\usepackage{paralist}
\usepackage[toc,page]{appendix}

\newcommand{\NAME}{TIFA\xspace}

\usepackage[pagebackref=true,breaklinks=true,letterpaper=true,colorlinks,bookmarks=false]{hyperref}

\iccvfinalcopy 

\def\iccvPaperID{9701} 

\ificcvfinal\fi

\begin{document}

\title{\NAME: Accurate and Interpretable Text-to-Image Faithfulness \\ Evaluation with Question Answering}

\author{
     \textbf{Yushi Hu}$^{1}$ \quad
     \textbf{Benlin Liu}$^{1}$ \quad
     \textbf{Jungo Kasai}$^{1}$ \quad
     \textbf{Yizhong Wang}$^{1}$ \quad \\
     \textbf{Mari Ostendorf}$^{1}$ \quad
     \textbf{Ranjay Krishna}$^{1,2}$ \quad
     \textbf{Noah A. Smith}$^{1,2}$ \\
     $^1$University of Washington\quad
     $^2$Allen Institute for AI\\
     {\tt \textcolor{pink}{\url{https://tifa-benchmark.github.io/}}}
}

\maketitle
\ificcvfinal\thispagestyle{empty}\fi

\begin{abstract}
Despite thousands of researchers, engineers, and artists actively working on improving text-to-image generation models, systems often fail to produce images that accurately align with the text inputs. 
We introduce \NAME (\textbf{T}ext-to-\textbf{I}mage \textbf{F}aithfulness evaluation with question \textbf{A}nswering), an automatic evaluation metric that measures the faithfulness of a generated image to its text input via visual question answering (VQA).
Specifically, given a text input, we automatically generate several question-answer pairs using a language model.
We calculate image faithfulness by checking whether existing VQA models can answer these questions using the generated image.
\NAME is a \emph{reference-free} metric that allows for fine-grained and interpretable evaluations of generated images.
\NAME also has better correlations with human judgments than existing metrics.
Based on this approach, we introduce \NAME v1.0, a benchmark consisting of 4K diverse text inputs and 25K questions across 12 categories (object,  counting, etc.).
We present a comprehensive evaluation of existing text-to-image models using \NAME v1.0 and highlight the limitations and challenges of current models.
For instance, we find that current text-to-image models, despite doing well on color and material, 
still struggle in counting, spatial relations, and composing multiple objects. 
We hope our benchmark will help carefully measure the research progress in text-to-image synthesis and provide valuable insights for further research.\footnote{Correspondance to <Yushi Hu: \url{yushihu@uw.edu}>.
All data and a pip-installable evaluation package are available on the project page.}
\end{abstract}

\begin{figure}[ht]
\centering
  \includegraphics[width=0.45\textwidth]{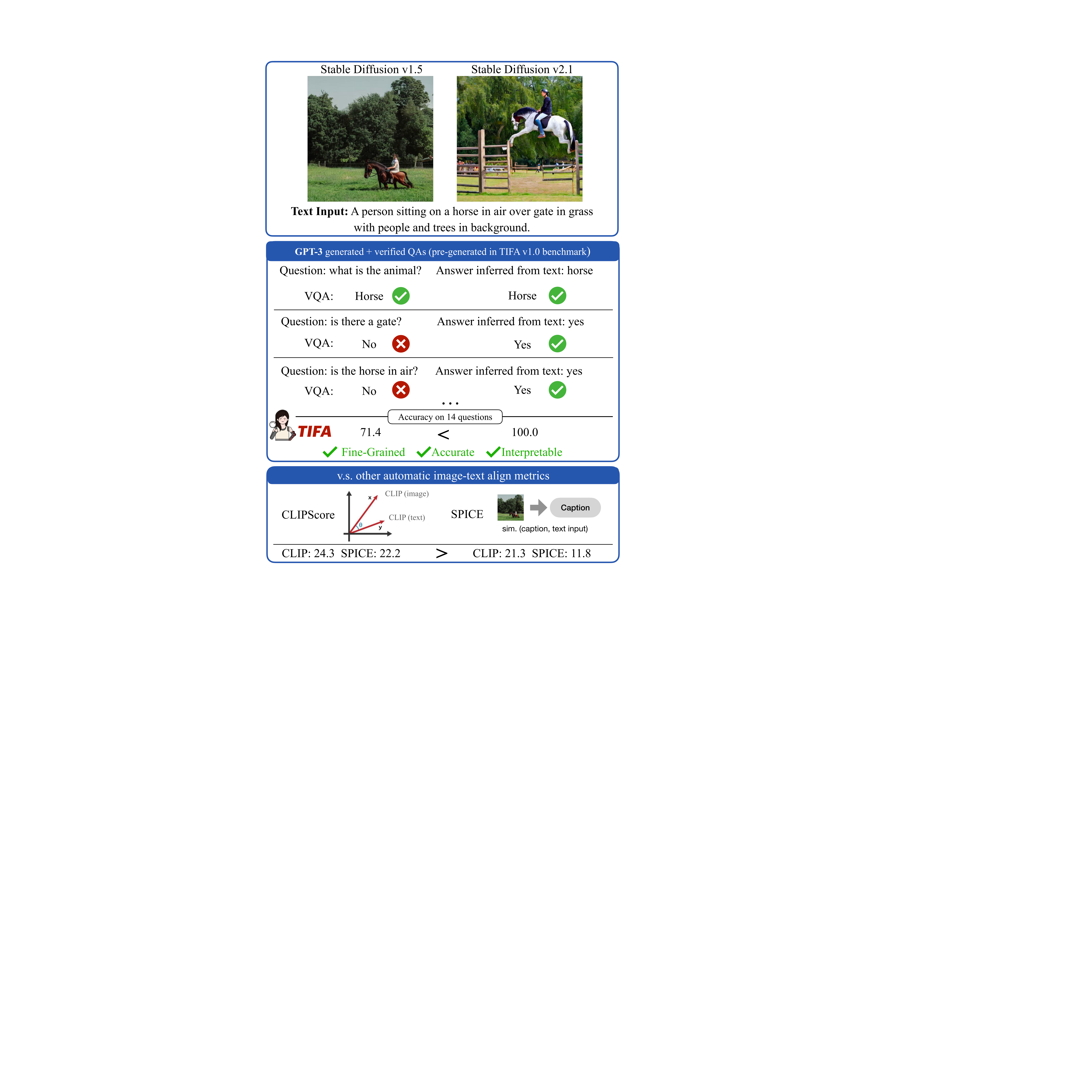}
  \caption{
Illustration of how \NAME works, and comparison with the widely-used CLIPScore and SPICE metrics.
Given the text input, \NAME uses GPT-3 to generate several question-answer pairs, and a QA model filters them (3 out of 14 questions for this text input are shown). 
\NAME measures whether VQA models can accurately answer these questions given the generated image.
In this example, \NAME indicates that the image generated by Stable Diffusion v2.1 is better than that by v1.5, while CLIP and SPICE yield the opposite result.
The text input is from the MSCOCO validation set.
}
  \vspace{-4mm}
  \label{fig:teaser}
\end{figure}

\begin{figure*}[t]
\centering
  \includegraphics[width=0.95\textwidth]{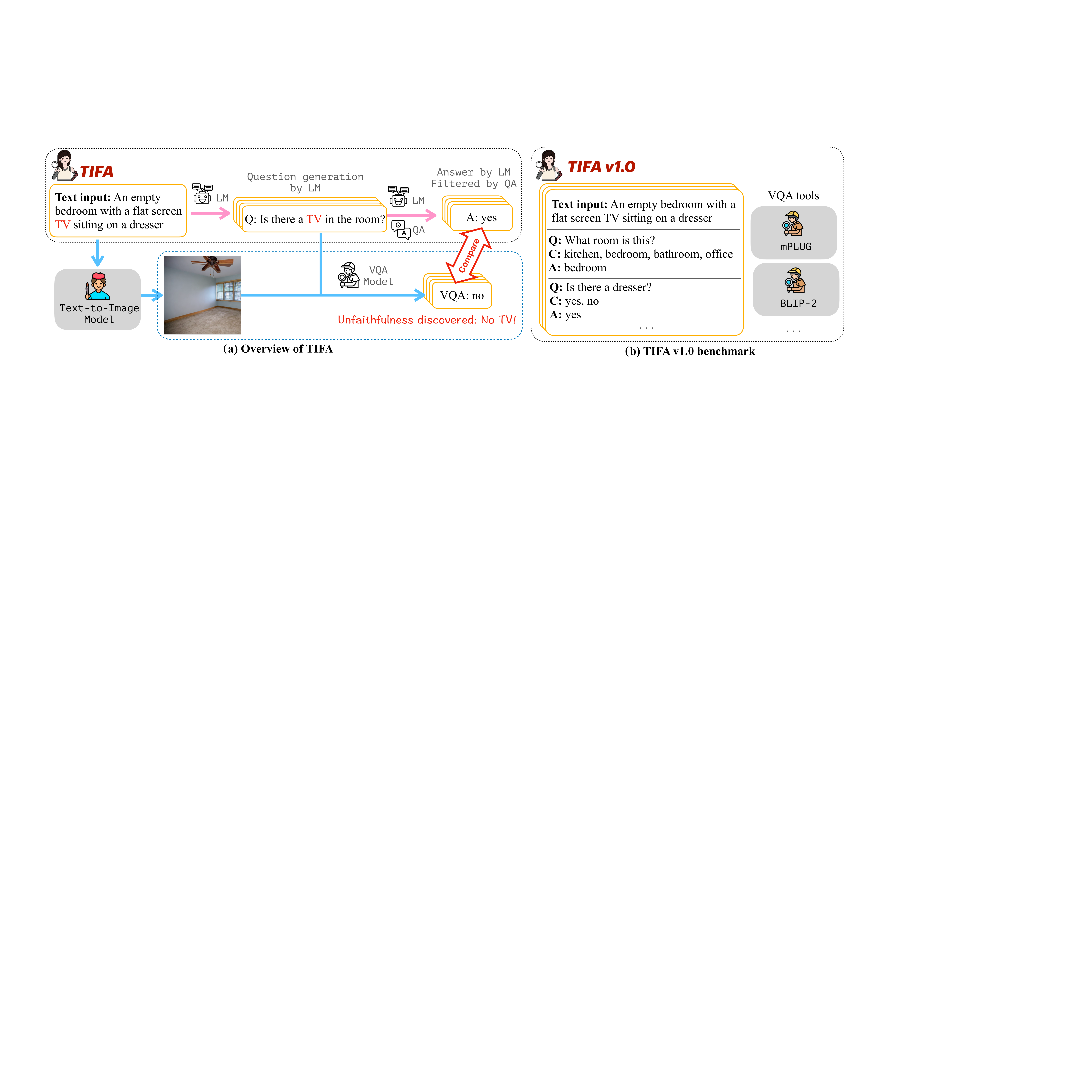}
  \caption{
\textbf{(a) Overview of how \NAME evaluates the faithfulness of a synthesized image.} \NAME uses a language model (LM), a question-answering (QA) model, and a visual-question-answering (VQA) model. Given a text input, we generate several question-answer pairs with the LM and then filter them via the QA model. To evaluate the faithfulness of a synthesized image to the text input, a VQA model answers these visual questions using the image, and we check the answers for correctness.
\textbf{(b) \NAME v1.0 benchmark.} While \NAME is applicable to any text prompt, to allow direct comparison across different studies, and for ease of use, we introduce the \NAME v1.0 benchmark, a repository of text inputs along with pre-generated question-answer tuples with answer choices. 
To evaluate a text-to-image model, a user first produces the images for the text inputs in \NAME v1.0 and then performs VQA with our provided tools on generated images to compute \NAME.
}
  \vspace{-4mm}
  \label{fig:overview}
\end{figure*}

\vspace{-0.2in}
\section{Introduction}
\label{sec:intro}

While we welcome artistic freedom when we commission art from artists, images produced by deep generative models~\cite{Ramesh2021ZeroShotTG, Rombach2021HighResolutionIS, Ramesh2022HierarchicalTI, Saharia2022PhotorealisticTD, Yu2022ScalingAM} should conform closely to our requests. 
Despite the advances in generative models, it is still challenging for models to produce images faithful to users' intentions~\cite{petsiuk2022human, Feng2022TrainingFreeSD, lee2023aligning, Liu2022CompositionalVG, Liu2022CharacterAwareMI}. 
For example, current models often fail to compose multiple objects~\cite{petsiuk2022human, Feng2022TrainingFreeSD, Liu2022CompositionalVG}, bind attributes to the wrong objects~\cite{Feng2022TrainingFreeSD}, and struggle in generating visual text~\cite{Liu2022CharacterAwareMI}.
Today, there are efforts to address these challenges: researchers are imposing linguistic structure with diffusion guidance to produce images with multiple objects~\cite{Feng2022TrainingFreeSD}; others are designing reward models trained using human feedback to better align generations with user intention~\cite{lee2023aligning}. 
However, progress is difficult to quantify without accurate and interpretable evaluation measures that explain when and how models struggle.

A critical bottleneck, therefore, is the lack of reliable automatic evaluation metrics for text-to-image generation faithfulness.
One of the popular metrics is CLIPScore~\cite{Hessel2021CLIPScoreAR}, which measures the cosine similarity between the CLIP embeddings~\cite{radford2021learning} of the text input and the generated image.
However, since CLIP is not effective at counting objects~\cite{radford2021learning}, or reasoning compositionally~\cite{Ma2022CREPECV}, CLIPScore is unreliable and often inaccurate.
Another family of evaluation metrics uses image captions, in which an image captioning model first converts the image into text, and then the image caption is evaluated by comparing it against the text input.
Unfortunately, using captioning models is insufficient since they might decide to ignore salient information in images or focus on other non-essential image regions~\cite{Kasai2021TransparentHE}; for example, a captioning model might say that the images in Figure~\ref{fig:teaser} are ``a field of grass with trees in the background''. Moreover, evaluating text (caption) generation is inherently challenging~\cite{Kasai2021billboard,khashabi-etal-2022-genie}.
Another recent text-to-image evaluation is DALL-Eval~\cite{Cho2022DALLEvalPT}, which employs object detection to determine if the objects in the texts are in the generated images. 
However, this approach only works on synthesized text and measures faithfulness along the limited axes of objects, counting, colors, and spatial relationships but misses activities, geolocation, weather, time, materials, shapes, sizes, and other potential categories we often ask about when we recall images from memory~\cite{Krishna2019InformationMV}.


To address the above challenges, we introduce \NAME, a new metric to evaluate text-to-image generation faithfulness. Our approach is illustrated in Figure~\ref{fig:overview}. Given a repository of text inputs, we automatically generate question-answer pairs for each text via a language model (here, GPT-3~\cite{brown2020language}). 
A question-answering (QA) system (here, UnifiedQA~\cite{Khashabi2020UnifiedQACF}) is subsequently used to verify and filter these question-answer pairs.
To evaluate a generated image, we use a visual-question-answering (VQA) system (here, mPLUG-large~\cite{Li2022mPLUGEA}, BLIP-2~\cite{Li2023BLIP2BL}, etc.) to answer the questions given the generated image.
We measure the image's faithfulness to the text input as the accuracy of the answers generated by the VQA system. 
While the accuracy of \NAME is dependent on the accuracy of the VQA model, our experiments show that \NAME has much higher correlation with human judgments than CLIPScore (Spearman's $\rho =$ 0.60 vs. 0.33) and captioning-based approaches (Spearman's $\rho =$ 0.60 vs. 0.34).
Additionally, since the LMs and VQA models will continue to improve, we hypothesize that \NAME will continue to be more reliable over time.
Also, our metrics can automatically detect when elements are missing in the generation: in Figure~\ref{fig:overview}, \NAME detects that the generated image does not contain a TV.

To promote the use of our new evaluation metric, we release \NAME v1.0,  
a large-scale text-to-image generation benchmark containing 4K diverse text inputs, sampled from the MSCOCO captions~\cite{lin2014microsoft}, DrawBench~\cite{Saharia2022PhotorealisticTD}, PartiPrompts~\cite{Yu2022ScalingAM}, and PaintSkill~\cite{Cho2022DALLEvalPT}. 
Each input comes with a pre-generated set of question-answer pairs, resulting in 25K questions covering 4.5K distinct elements.
These questions have been automatically generated and pre-filtered using a question-answering model.
This benchmark also comes with different VQA models~\cite{wang2022git, kim2021vilt, wang2022ofa, Li2023BLIP2BL, Li2022mPLUGEA, Hu2022PromptCapPT} that can be used to evaluate generative models and can be easily extended to use future VQA models when they become available.

We conduct a comprehensive evaluation of current text-to-image models using \NAME v1.0.
Thanks to \NAME's ability to detect fine-grained unfaithfulness in images, we find that current state-of-the-art models are good at rendering common objects, animals, and colors, but still struggle in composing multiple objects, reasoning about spatial relations, and binding the correct activity for each entity.
In addition, our ablation experiments show that \NAME is robust to different VQA models.
Future researchers can use \NAME v1.0 to compare their text-to-image models' faithfulness across different studies. 
Also, future generative models may focus on addressing the weaknesses of current models that \NAME discovered.
In addition, with \NAME, users can customize evaluations with their own text inputs and questions~\cite{ethayarajh-jurafsky-2020-utility}; for example, a future \NAME benchmark could focus on counting or scene text.

\section{Related Work}
\label{sec:related}

We compare \NAME to other image and language generation evaluation metrics.

\noindent\textbf{Prior image generation evaluation}
Prior work usually compares image generation models via pairwise comparison by humans. How to design automatic evaluation metrics to approximate human assessment of the quality of machine-generated images has always been a major challenge in computer vision. There are two aspects to evaluate, namely image quality and image-text alignment.~\textbf{Inception Score}~\cite{Salimans2016ImprovedTF} and \textbf{FID}~\cite{Heusel2017GANsTB} are the most widely adopted metrics for image quality. They compare the features of the generated images and gold images extracted from a pre-trained Inception-V3 model~\cite{Szegedy2015RethinkingTI} to evaluate the fidelity and diversity of generated images. However, they rely on ground-truth images and are based on a classification model, which makes them not suitable for complex datasets~\cite{Frolov2021AdversarialTS}. For image-text alignment, prior metrics are mainly based on CLIP~\cite{radford2021learning}, captioning, and object detection models. \textbf{CLIPScore}~\cite{Hessel2021CLIPScoreAR} and \textbf{CLIP-R}~\cite{park2021benchmark} are based on the cosine similarity of image and text CLIP~\cite{radford2021learning} embeddings. \cite{Cho2022DALLEvalPT, Hinz2019SemanticOA, Hong2018InferringSL} first convert the images using a captioning model, and then compare the image caption with the text using metrics like CIDEr~\cite{vedantam2015cider} and SPICE~\cite{anderson2016spice}. \textbf{SOA}~\cite{Hinz2019SemanticOA} and \textbf{DALL-Eval}~\cite{Cho2022DALLEvalPT} employ object detection models to determine if objects, attributes, and relations in the text input are in the generated image. However, this approach only works on synthesized text inputs and measures faithfulness on limited axes (object, counting, color, spatial relation), missing elements like material, shape, activities, and context. In contrast, thanks to the flexibility of questions, \NAME works on any text inputs and evaluates faithfulness across a broad spectrum of dimensions.

\noindent\textbf{Summarization evaluation in NLP}
\NAME is inspired by the summarization evaluation methods based on question answering (QA) \cite{Wang2020AskingAA, singh2019towards}. Given a summary, a language model generates a set of questions about the text. A QA model checks if the same answer can be inferred from the text and the summary. These QA-based metrics have much higher correlations with human judgments on the factual consistency of summarization than other automatic metrics \cite{Wang2020AskingAA, singh2019towards}.
\NAME can be seen as treating the text input for the text-to-image model as a summary of the generated image.


\section{The \NAME Metric}
\label{sec:method}

\begin{figure*}[t]
\centering
  \includegraphics[width=0.75\textwidth]{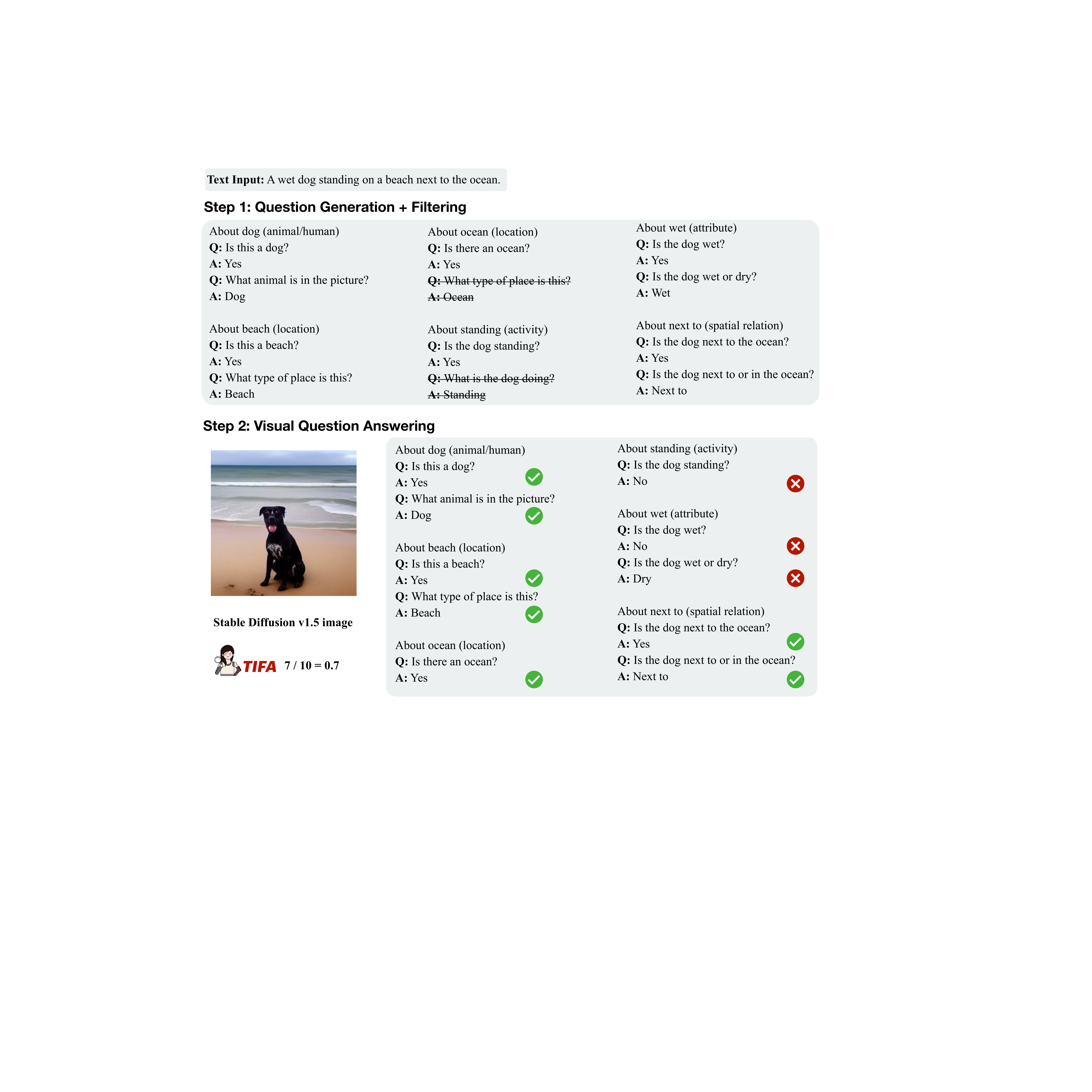}
  \caption{
Step-by-step qualitative example of \NAME metric. Given a text input, we first generate question-answer pairs and filter them. We \st{strikethrough} the questions filtered out by UnifiedQA. Then we run VQA models on the generated image to get the \NAME score. 
}

  \vspace{-4mm}
  \label{fig:step-example}
  
\end{figure*}

We introduce a framework for automatically estimating the faithfulness of an image to its text prompt.
Given a text input $T$, we aim to measure the faithfulness of the generated image $I$.
An overview of our metric is illustrated in Figure~\ref{fig:overview}.
From $T$, we generate $N$ multiple-choice question-answer tuples $\{Q_i, C_i, A_i\}_{i=1}^N$, in which $Q_i$ is a question, $C_i$ is a set of answer choices, and $A_i \in C_i$ is the gold answer. The answer $A_i$ can be inferred given $T$, $Q_i$, and $C_i$.
Next, for each question $Q_i$, we use a VQA model to produce an answer $A_i^{\text{VQA}} = \max_{a \in C_i} p(a \mid I, Q_i)$.
We define the faithfulness between the text $T$ and image $I$ as the VQA accuracy:
\begin{equation}
\vspace{-0.1in}
    \mathit{faithfulness}(T, I) = \frac{1}{N}\sum_{i=1}^N \mathbbm{1}[A_i^{\text{VQA}} = A_i]
\end{equation}
The range of our faithfulness score is $[0,1]$. It is maximized when we have a performant VQA model, and the image $I$ accurately covers the information in the text $T$ so that for any question $Q$, which can be answered given $T$ can also be answered given $I$.
Several key design decisions will be addressed in later sections: how to generate questions (\S\ref{sec:method:qg}), how to control the question quality (\S\ref{sec:filtering}), and how to answer those questions (\S\ref{sec:answering}). Finally, we give a step-by-step qualitative example of \NAME in Figure~\ref{fig:step-example}.




\subsection{Question-Answer Generation}
\label{sec:method:qg}

Our main challenge is to generate diverse questions that cover all elements of the text input evenly. We also simplify the question-generation pipeline into a single GPT-3~\cite{brown2020language} completion, so that \NAME can exploit the power of recent language models (LM) and work with updated black-box LMs (e.g., ChatGPT) in the future.

Inspired by prior work~\cite{Changpinyo2022AllYM}, given a text prompt $T$, we generate the question-answer tuples $\{Q_i, C_i, A_i\}_{i=1}^N$ via the pipeline illustrated in Figure~\ref{fig:question_generation}.
Different from prior work, which relies on multiple components, our pipeline is completed by a single inference run via in-context learning with GPT-3 \cite{brown2020language, wei2022chain, hu2022incontext, press2022measuring, su-etal-2023}, thereby avoiding the need for intermediate human annotations.
We annotate 15 examples and use them as in-context examples for GPT-3 to follow.
Here we take the text \emph{``A photo of three dogs.''} as an example.
Each in-context example contains the following steps:

\begin{figure}[t]
\centering
  \includegraphics[width=0.42\textwidth]{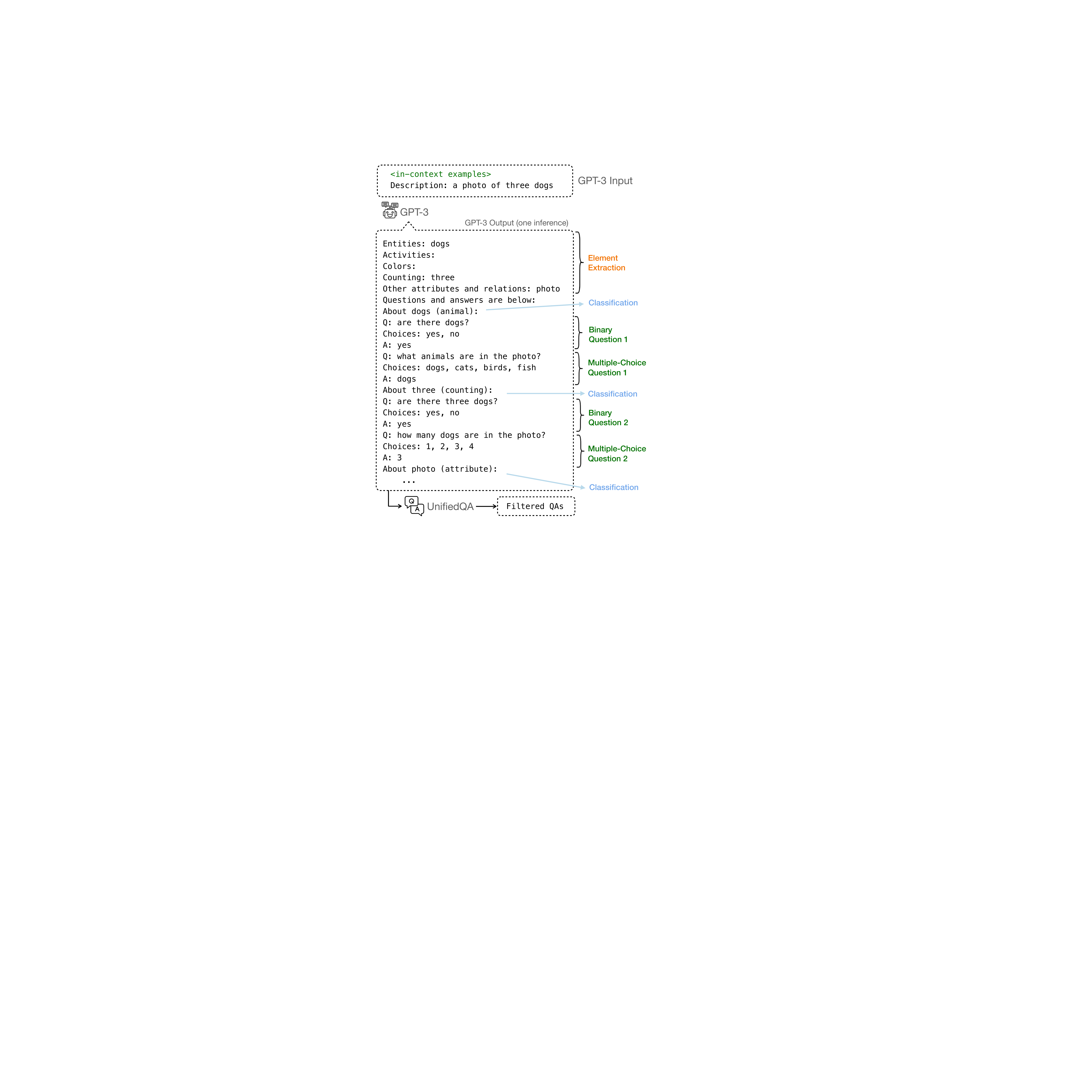}
  \caption{
Our question-answer pair generation pipeline. The whole pipeline can be executed via a single inference of GPT-3 via in-context learning. Given the text prompt, GPT-3 first extracts the elements and then generates two questions for each element. The GPT-3 output is then parsed and filtered by UnifiedQA.
} 
  \vspace{-4mm}
  \label{fig:question_generation}
\end{figure}

\vspace{-0.15in}
\paragraph{Element extraction}
Given text prompt $T$, GPT-3  will first extract all elements $\{v_i\}_{i=1}^m$ following prior work~\cite{Changpinyo2022AllYM} (for the in-context examples, we perform element extraction manually).
The elements include noun phrases (including named entities), verbs, adjectives, adverbs, and parse tree spans with no more than 3 words altogether. For the above example, the elements are \emph{photo, three, dogs}.

\vspace{-0.15in}
\paragraph{Element category classification} 
For each element $v_i$, following~\cite{Krishna2019InformationMV}, we classify the elements into one of the following 12 categories: \emph{object, activity, animal, food, counting, color, material, spatial, location, shape, attribute}, and {other}. 
As shown in Figure~\ref{fig:question_generation}, text generated from GPT-3 contains the category corresponding to each question.
For example, ``three'' is ``counting'', and ``dogs'' is classified as ``animal.''
This step allows a detailed analysis of the text-to-image model's ability in each category.

\vspace{-0.15in}
\paragraph{Question generation conditioned on elements}
For each element $v_i$, we generate two questions. The first is a question that should be answered ``yes'' for a faithful generated image, and the second question has $v_i$ as its answer. For example, two questions are generated for the element \emph{``three''}. The first is ``are there three dogs?'', and the choices are \{yes, no\}. The second is ``how many dogs are there?'', and the choices are \{1,2,3,4\}.
These two types of questions make our evaluations diverse and robust to surface-level differences.

\vspace{-0.15in}
\paragraph{Completing above steps by prompting GPT-3 once}
As mentioned earlier, for each text $T$, the whole pipeline can be completed by one GPT-3 inference. 
We annotated 15 in-context examples that cover all types of questions. The full prompt is in the Appendix.
The prompt format is shown in Figure~\ref{fig:question_generation}. 
Our in-context examples follow the same format, and identical examples are used for all text inputs, leading to a fixed and limited amount of human annotation cost.
We use \textit{code-davinci-002} engine for question generation, and the decoding temperature is 0.

\subsection{Question Filtering}
\label{sec:filtering}
To ensure the quality of generated images, we use UnifiedQA~\cite{Khashabi2020UnifiedQACF} to verify the GPT-3 generated question-answer pairs and filter out the ones that GPT-3 and UnifiedQA do not agree on.
UnifiedQA\footnote{Model checkpoint we use: \url{https://huggingface.co/allenai/unifiedqa-v2-t5-large-1363200}.} is a state-of-the-art multi-task question-answering model that can answer both multiple-choice and free-form questions. 
Denote the UnifiedQA model as $\mathit{QA}$. Given the text $T$, question $Q_i$, choices $C_i$, and answer $A_i$, Let $A_i^{f} = QA(T, Q_i)$ be the free-form answer, and $A_i^{mc} = \mathit{QA}(T, Q_i, C_i)$ be the multiple-choice answer.
We keep the question if $A_i = A_i^{mc}$ and the word-level $F_1$ score between $A_i^{f}$ and $A_i$ is greater than 0.7.
We conduct a human evaluation on 1000 filtered question-answer pairs.
Only 7 are considered not reasonable (e.g., generated choices do not include a correct answer).
Details are in Appendix~\ref{appendix:annotation}.

\vspace{-0.15in}
\paragraph{Recommended VQA model} Based on considerations over the accuracy, correlation with human judgments, and run time, we currently suggest using \textbf{mPLUG-large} as the VQA model for \NAME. Analysis is given in Section~\ref{sec:exp:comparisonVQA}.
Like the LM and QA components, the VQA component can be updated in the future as the technology improves.

\subsection{VQA Models}
\label{sec:answering}
Since our questions contain a diverse set of visual elements (e.g., activity, art style), we use open-domain pre-trained vision-language models as our VQA model (rather than closed-class classification models fine-tuned on VQAv2~\cite{goyal2017making}). We provide tools to easily perform VQA on arbitrary images and questions, based on 5 state-of-the-art VQA models trained with distinct data and strategies. 

\vspace{-0.15in}
\paragraph{Vision-language models}
The general pre-trained vision-language model are \textbf{GIT-large}~\cite{wang2022git}, \textbf{VILT-B/32}~\cite{kim2021vilt}, \textbf{OFA-large}~\cite{wang2022ofa}, and \textbf{mPLUG-large}~\cite{Li2022mPLUGEA}. 
These models are pre-trained on a large amount of image-text pairs, and downstream image-to-text tasks like image captioning and visual question answering.
Notice that these models have not been trained to answer multiple-choice questions. For each question, we first decode the free-form answer and then choose the choice that has the highest similarity with the decoded answer, measured by SBERT~\cite{Reimers2019SentenceBERTSE}.
Another model we use is \textbf{BLIP-2 FlanT5-XL}~\cite{Li2023BLIP2BL}, in which a VIT~\cite{dosovitskiy2021an} is connected with a frozen FlanT5~\cite{Chung2022ScalingIL} via a lightweight transformer. This model allows for performing multiple-choice VQA directly due to the flexibility of the LM.


\section{\NAME v1.0: Benchmark for Text-to-Image Generation Faithfulness}
\label{sec:benchmark}


In this section, we introduce \NAME v1.0, a text-to-image generation faithfulness benchmark based on the evaluation method discussed in Section~\ref{sec:method}. The benchmark consists of 4,081 diverse text inputs paired with 25,829 question-answer pairs. Each question is classified into one of the categories discussed in Section~\ref{sec:method:qg}. The benchmark also comes with Python pip-installable APIs to perform VQA with various state-of-the-art VQA models on arbitrary visual questions.
The overall \NAME for each text-to-image model is computed by averaging \NAME scores of images generated from each text input in the benchmark.

\subsection{Text Collections}
We collect 4,081 text inputs to benchmark text-to-image models' generation ability on diverse tasks.
2,000 text inputs are image captions from \textbf{COCO} validation set~\cite{lin2014microsoft}.
These captions have corresponding gold images. 
Since text-to-image models are often used to create abstract art, we also collect 2,081 text inputs from previous works that do not correspond to any real image.
All text inputs we use contain $\ge 3$ words.
We include 161 from \textbf{DrawBench} used in Imagen~\cite{Saharia2022PhotorealisticTD} (texts that are categorized as ``misspellings'' and ``rare words'' are removed); 1420 from \textbf{PartiPrompt} used in Parti~\cite{Yu2022ScalingAM} (texts in category ``abstract'' are removed); and 500 texts from \textbf{PaintSkill} used in DALL-Eval~\cite{Cho2022DALLEvalPT}.

\begin{figure*}[t]
\centering
  \includegraphics[width=0.98\textwidth]{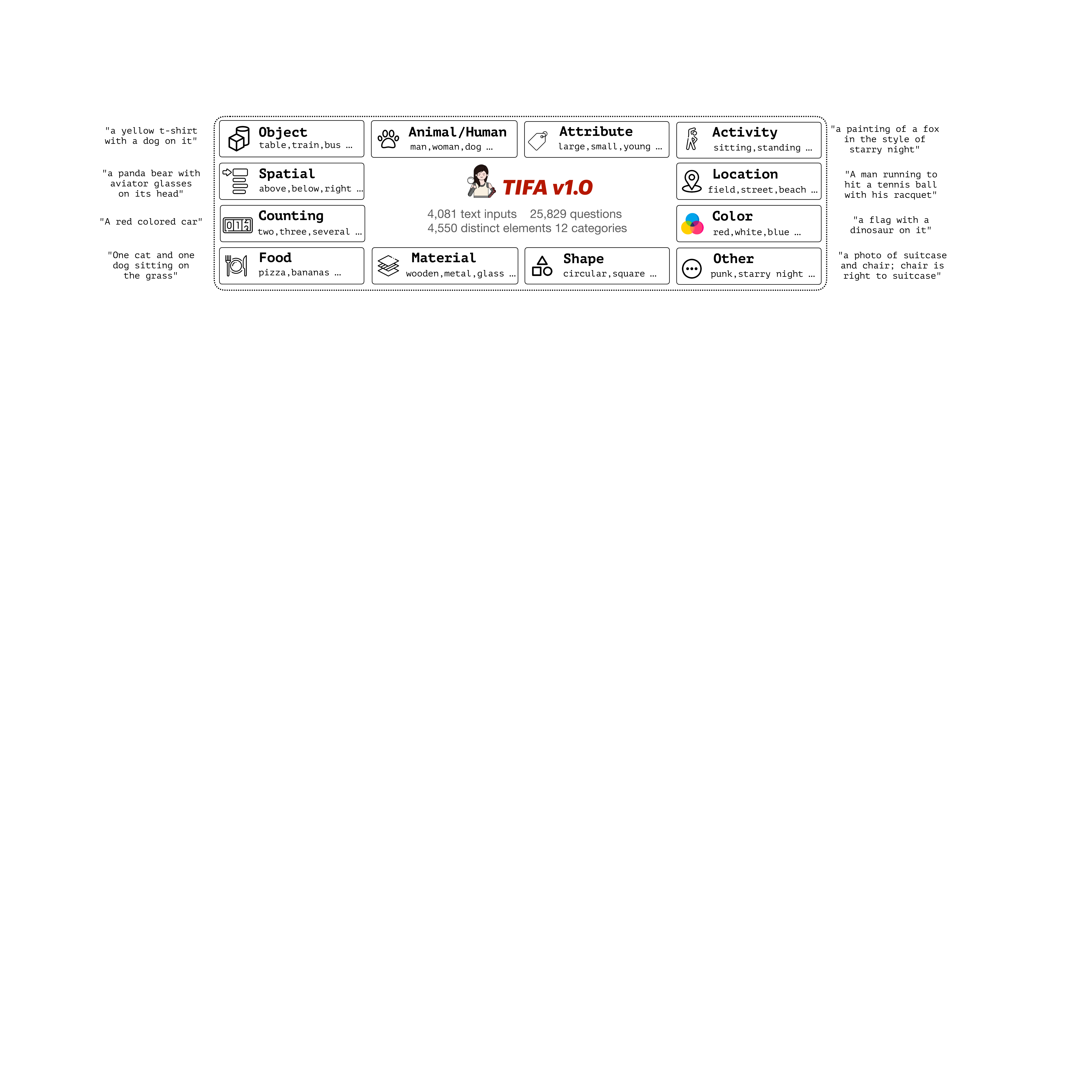}
  \caption{
Statistics and diversity of \NAME v1.0. The text inputs contain elements from 12 categories (e.g., object, spatial, and counting). We show the most common elements from each category. In addtion, we also show some example text inputs on the sides.
}
  \vspace{-4mm}
  \label{fig:benchmark}
  
\end{figure*}

\begin{table}[h]
\small
\centering
\caption{
Statistics of \NAME v1.0.
}
\begin{tabular}{lr}
\toprule[1.2pt]
Statistics \\
\midrule
\# of prompts & 4,081 \\
$\ $ - \# of COCO captions & 2,000 \\
$\ $ - \# of DrawBench, PartiPrompt, PaintSkill prompts & 2,081 \\
\midrule
\# of questions & 25,829 \\
$\ $ - \# of binary questions & 17,226 \\
$\ $ - \# of multiple-choice questions & 8,603 \\
\midrule
avg. \# of questions per prompt & 6.3 \\
avg. \# of words per prompt & 10.5 \\
avg. \# of elements per prompt & 4.3 \\
\bottomrule[1.2pt]
\end{tabular}
\label{tab:statistics}
\vspace{-3mm}
\end{table}

\subsection{Statistics and Diversity}
Table~\ref{tab:statistics} shows the basic statistics of the \NAME v1.0 benchmark.
We demonstrate \NAME v1.0's diversity in Figure~\ref{fig:benchmark}.
\NAME v1.0 contains questions about 4,550 distinct elements, which are categorized into 12 categories. 
The number of times each type of element occurs in the text input is object (7,854), animal/human (3,501), attribute (3,399), activity (2,851), spatial (2,265), location (1,840), color (1,743), counting (986), food (911), material (209), shape (69), and other (201).
The "other" category includes notions often used in abstract art, such as ``starry night'' and ``steampunk.''
The accuracy of VQA on a particular genre measures the text-to-image model's ability in the corresponding aspect.
Please refer to Appendix~\ref{appendix:qa} for more details on \NAME v1.0.


  

\subsection{Finetuned Open-Source Language Model for Question Generation}
\label{sec:question_generation_llama}
For \NAME v1.0, we use GPT-3 to generate the questions. While benchmarking with TIFA v1.0 is a deterministic process, using \NAME to create a new benchmark might not be deterministic as the underlying question generator (GPT-3 in our case) might change privately.
To promote deterministic benchmark generation, we fine-tune and release a LLaMA 2 (7B)~\cite{touvron2023llama} model that parses the captions and generates questions for arbitrary texts, using \NAME v1.0 questions as training examples.
\footnote{LLaMA 2 question generation model checkpoint: \\ \url{https://huggingface.co/tifa-benchmark/llama2_tifa_question_generation}}

\section{Experiments}
\label{sec:exp}

In this section, we first show that \NAME has substantially higher correlations with human judgments than prior metrics on text-to-image faithfulness (\S\ref{sec:exp:corr}). Then we present a comprehensive evaluation of existing text-to-image models using \NAME v1.0, highlighting the challenges of current text-to-image models (\S\ref{sec:exp:findings}). Finally, we conduct an analysis of \NAME's robustness against different VQA models (\S\ref{sec:exp:comparisonVQA}). For all experiments, we use \textbf{mPLUG} as the VQA model for \NAME unless stated otherwise.
The models we evaluate include AttnGAN~\cite{Xu2017AttnGANFT}, X-LXMERT~\cite{Cho2020XLXMERTPC}, VQ-Diffusion~\cite{Gu2021VectorQD}, minDALL-E~\cite{kakaobrain2021minDALL-E}, and Stable Diffusion v1.1, v1.5, and v2.1~\cite{Rombach2021HighResolutionIS}. Details are in Appendix~\ref{appendix:t2i}.

\subsection{Correlation with Human Judgements}
\label{sec:exp:corr}

To compare \NAME with prior evaluation metrics, we first conduct human evaluations of the text-to-image models on the 1-5 Likert scale on text-to-image faithfulness.
Then we compare \NAME with other metrics based on their correlation with human judgments.

\begin{figure}[h]
\centering
  \includegraphics[width=0.35\textwidth]{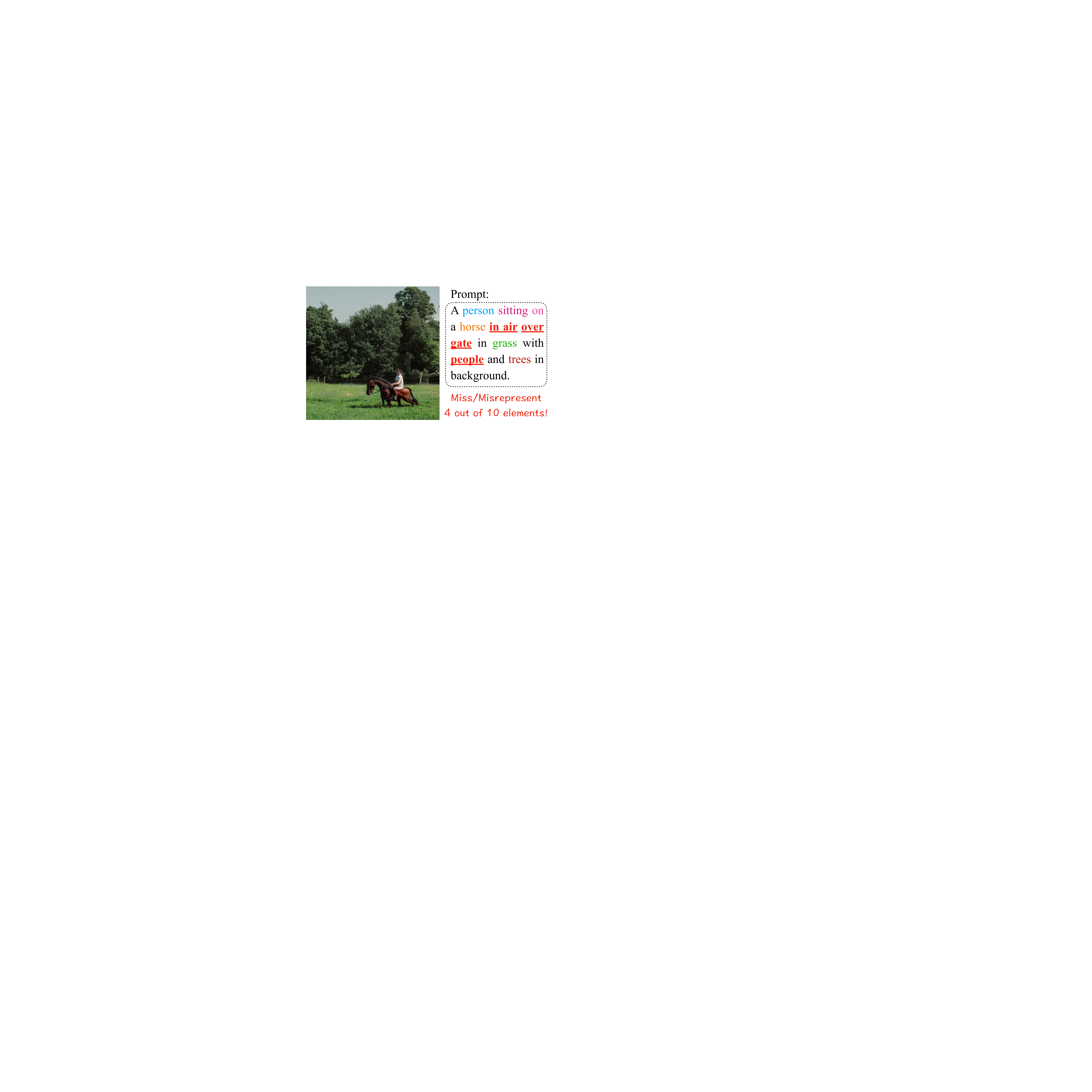}
  \caption{
Illustration of our Likert scale annotation guideline. Annotators are asked to give a score of 1 to 5 based on how many elements in the text prompt are missed or misrepresented in the image. 
The missed elements are \textcolor{red}{\underline{underlined}}.
}
  \label{fig:human_eval}
\end{figure}

\vspace{0.1in}\noindent\textbf{Likert scale on text-to-image faithfulness}
Annotators are asked to answer on a scale of 1 (worst) to 5 (best) to the question \emph{``Does the image match the text?"}.
The detailed annotation guidelines are in Appendix~\ref{appendix:annotation}.
Annotators are asked to focus on text-to-image faithfulness rather than image quality.
The Likert scale should be based on how many elements in the text prompt are missed or misrepresented in the image.
Objects are more important than attributes, relations, and activities.
If an object is missed in the image, then all related attributes, activities, relations, etc. are also considered lost.
An example is given in Figure~\ref{fig:human_eval}. 

We collect annotations of 800 generated images on 160 text inputs from \NAME v1.0. For each prompt, we sample an image from the 5 most recent generative models we evaluated, i.e., minDALL-E, VQ-Diffusion, Stable Diffusion v1.1, v1.5, and v2.1.  We collect 2 annotations per image and average over the scores as the single ``faithfulness" score. The inter-annotator agreement measured by Krippendorf's $\alpha$ is 0.67, indicating ``substantial" agreement.

\vspace{0.1in}\noindent\textbf{Baselines} 
We compare our evaluation with two families of reference-free metrics on text-image match introduced in Section~\ref{sec:related}. The first is the \textbf{caption-based method}.
We use the state-of-the-art \textbf{BLIP-2 FlanT5-XL}~\cite{Li2023BLIP2BL} as the captioning model.
The second approach is \textbf{CLIPScore}~\cite{Hessel2021CLIPScoreAR, radford2021learning}. We use CLIP (VIT-B/32)~\cite{radford2021learning} to compute the score.

\begin{table}[h]
\small
\centering
\caption{
Correlations between each evaluation metric and human judgment on text-to-image faithfulness, measured by Spearman's $\rho$ and Kendall's $\tau$. 
}
\begin{tabular}{lcc}
\toprule[1.2pt]
 & Spearman's $\rho$ & Kendall's $\tau$ \\
\midrule
\textbf{Caption-Based} \\
BLEU-4 & 18.3 & 18.8\\
ROUGE-L & 32.9 & 24.5\\
METEOR & 34.0 & 27.4\\
SPICE & 32.8 & 23.2\\
\midrule
CLIPScore & 33.2 & 23.1\\
\midrule
\textbf{Ours} \\
\NAME (VILT) & 49.3 & 38.2\\
\NAME (OFA) & 49.6 & 37.2\\
\NAME (GIT) & 54.5 & 42.6\\
\NAME (BLIP-2) & 55.9 & 43.6\\
\textbf{\NAME (mPLUG)} & \textbf{59.7} & \textbf{47.2}\\
\bottomrule[1.2pt]
\end{tabular}
\label{tab:human_correlation}
\vspace{-0.1in}
\end{table}

\vspace{0.1in}\noindent\textbf{\NAME has a much higher correlation with human judgments than prior metrics.}
The correlations between each evaluation metric and human judgment are shown in Table~\ref{tab:human_correlation}. For caption-based evaluations, we use metrics 
BLEU-4~\cite{papineni2002bleu}, ROUGE-L~\cite{lin-2004-rouge}, METEOR~\cite{banerjee2005meteor}, and SPICE~\cite{anderson2016spice}.
\NAME has higher correlations with human judgments than all previous evaluation metrics on all VQA models.
\NAME(mPLUG) yields the highest correlation with human judgments among all VQA models. 

\subsection{Benchmarking Text-to-Image Models}

\begin{figure}[th]
\centering
  \includegraphics[width=0.45\textwidth]{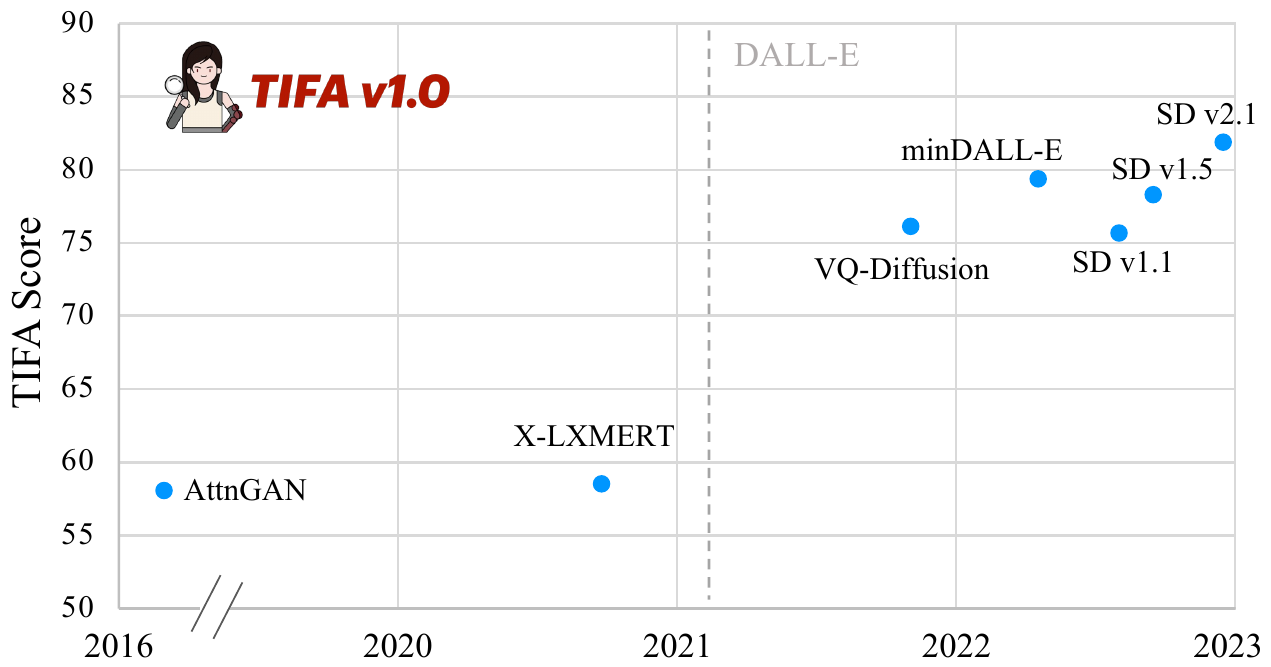}
  \caption{
Average \NAME score of text-to-image models on the \NAME v1.0 benchmark.
The horizontal axis shows their release dates.
}
  \vspace{-4mm}
  \label{fig:leaderboard}
\end{figure}

Figure~\ref{fig:leaderboard} shows the average \NAME score text-to-image models get on \NAME v1.0. The detailed scores with each VQA model on each element type are provided in Appendix~\ref{appendix:results}. We can see a clear trend of how text-to-image models evolve over time. There is a jump in \NAME score after DALL-E~\cite{Ramesh2021ZeroShotTG} is released, from about 60\% to 75\%. Qualitative examples of our evaluation metric are in Appendix~\ref{appendix:example}.

\begin{figure*}[ht]
\centering
  \includegraphics[width=0.95\textwidth]{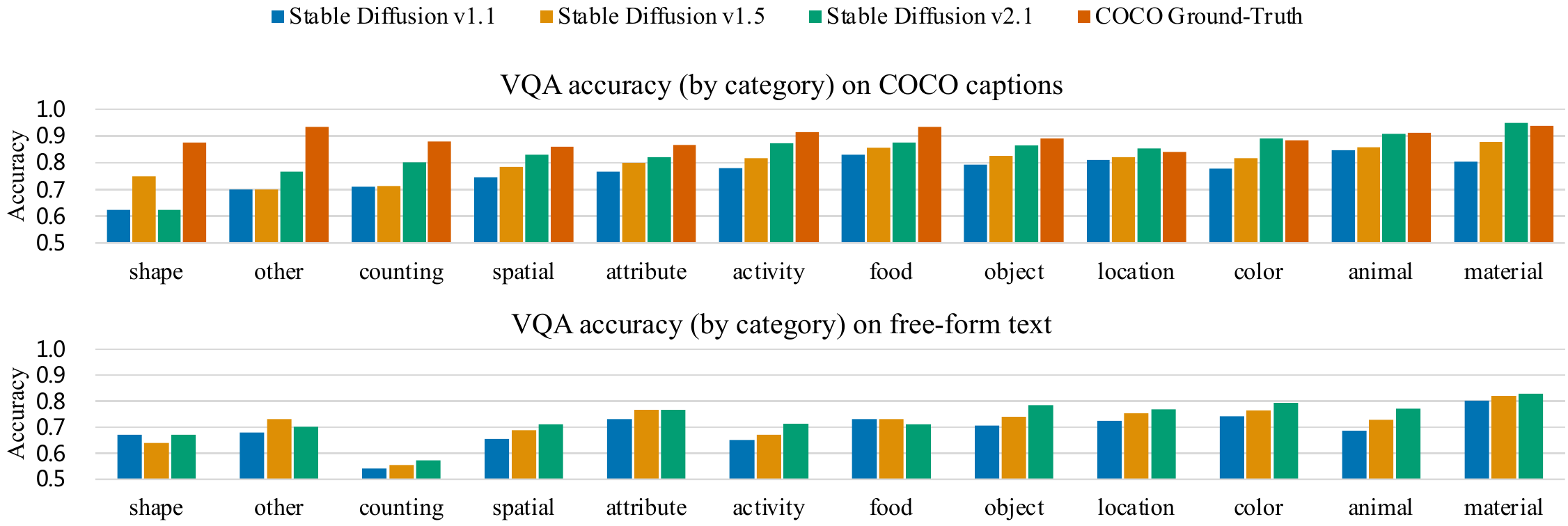}
  \caption{
 Accuracy on each type of question in the \NAME v1.0 benchmark. The text-to-image models are Stable Diffusion v1.1, v1.5, and v2.1. We order the categories by the average score Stable Diffusion v2.1 gets on corresponding questions. For COCO captions, we also include the accuracy of the ground-truth images for reference. 
}
  \label{fig:type}
\end{figure*}

\subsection{Findings on Current Text-to-Image Models}
\label{sec:exp:findings}

Figure~\ref{fig:type} shows accuracy on each type of question in \NAME v1.0 for Stable Diffusion v1.1, v1.5, and v2.1. The score on each type reflects the text-to-image models' faithfulness in each type of visual element.
To the best of our knowledge, \NAME is the only automatic evaluation method that can provide such a detailed fine-grained analysis of image generation. We separate the scores on COCO captions and other text inputs. For COCO captions, we also include the accuracy on the ground-truth images for reference. We summarize our findings in the following paragraphs.

\vspace{0.1in}\noindent\textbf{Generating images from captions vs.\ free-form text}
From Figure \ref{fig:type}, we can see that VQA accuracy is higher on the COCO captions than on other text inputs. The reason is that COCO captions correspond to real images, while other text inputs may correspond to compositions that cannot be found in real-world photos (e.g.~``a blue apple").

\vspace{0.1in}\noindent\textbf{What elements are text-to-image models struggling with?}
Based on the scores of each category in Figure~\ref{fig:type}, we can see that Stable Diffusion models are performing well on material, animal/human, color, and location in terms of text-to-image faithfulness.
However, they yield low accuracy on questions involving \textbf{shapes}, \textbf{counting}, and \textbf{spatial relations}. ``Other" mainly contains \textbf{abstract art notions}, and models are also struggling with them.
There is also a big gap between the synthesized images and real images on the COCO captions. Future work can explore various directions (e.g., training data/loss and model architecture)  to improve text-to-image models' faithfulness in these aspects.

\vspace{0.1in}\noindent\textbf{Why are ground-truth images not getting perfect scores?}
Ground-truth images in COCO do not get perfect scores because 1) the COCO captions contain a substantial amount of noise from crowd workers \cite{Kasai2021TransparentHE} and 2) VQA models are not perfect.
Real images have higher accuracy in all categories except material, color and location, where differences are small. It is left to future work to determine whether this is simply due to noise or it is an area where assessment can be improved. 

\vspace{0.1in}\noindent\textbf{Stable Diffusion is evolving.}
We can see the consistent trend that Stable Diffusion models are improving in their later versions in most of the element categories. The exceptions are ``shape" for both prompt sources, ``other" and ``food" for the free-form text inputs without gold images.

\begin{figure}[ht]
\centering
  \includegraphics[width=0.45\textwidth]{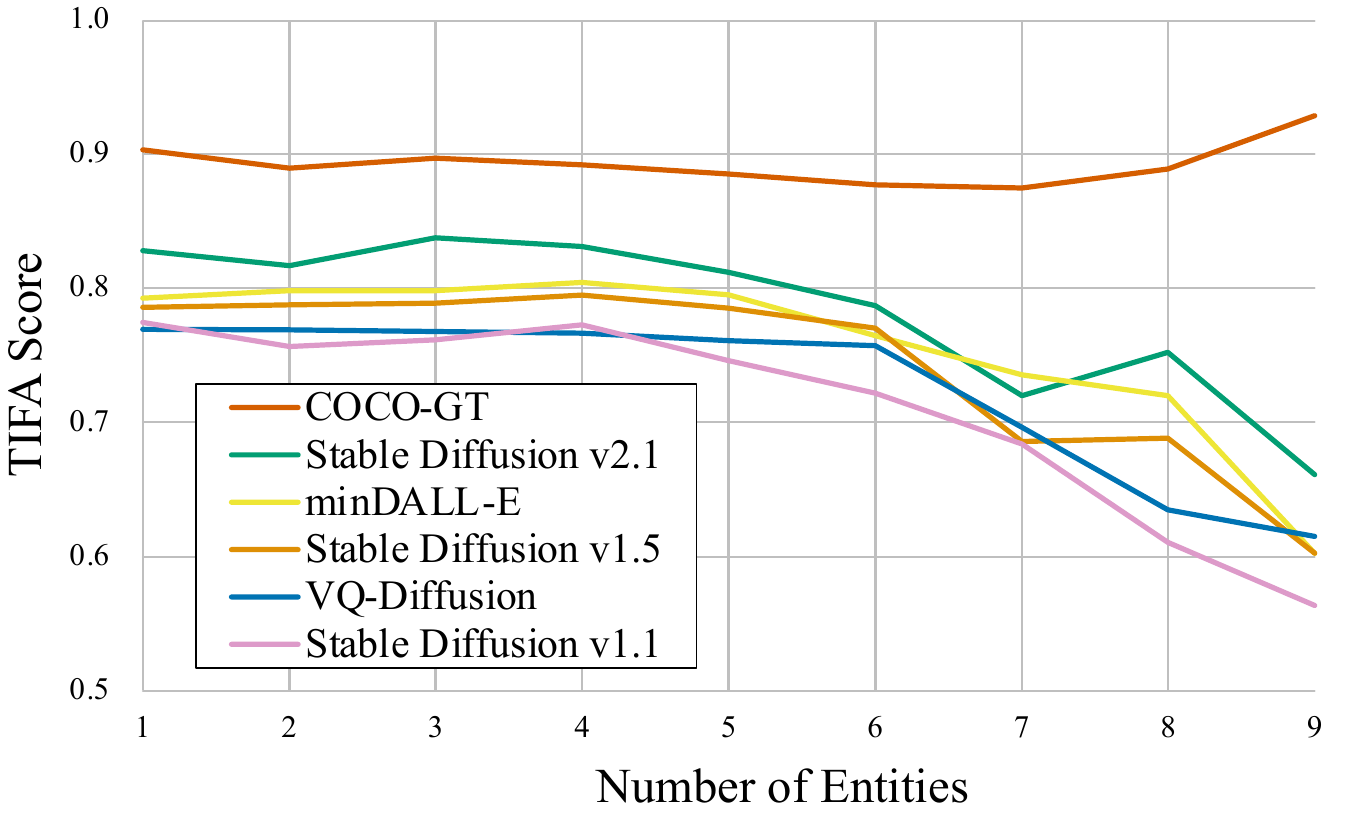}
  \caption{
\NAME vs.\ numbers of entities (objects, animals/humans, and food) in the text input. The accuracy starts to drop when more than 5 entities are added to the text, showing that compositionality is hard for text-to-image models. Meanwhile, \NAME scores for COCO ground-truth (GT) images remain consistent.
}

  \label{fig:compositionality}
\end{figure}

\vspace{0.1in}\noindent\textbf{Composing multiple objects is challenging.}
Figure~\ref{fig:compositionality} shows how the number of entities (objects, animals/humans, food) in the text input affects the average \NAME score.
When there are more than 5 entities, The \NAME score starts to drop rapidly for all text-to-image models, consistent with similar findings in other vision-language evaluations~\cite{grunde2021agqa,gandhi2022measuring}. For reference, we also add the real images in COCO in this figure. The \NAME score on real images is rather consistent and does not change as the number of entities increases. This quantitatively shows that composing multiple objects is challenging for current text-to-image models. 
One possible reason is that
the CLIP text embedding, which is used to train Stable Diffusion, lacks compositionality, as investigated in ~\cite{Ma2022CREPECV}.



\subsection{Analysis of VQA Models}
\label{sec:exp:comparisonVQA}
One major concern of \NAME is that VQA models can introduce some errors.
Table~\ref{tab:human_correlation} shows that \NAME has a much higher correlation with human judgment than the previous metrics, regardless of the choice of the VQA models; here we conduct a more detailed analysis.

\begin{figure}[ht]
\centering
  \includegraphics[width=0.45\textwidth]{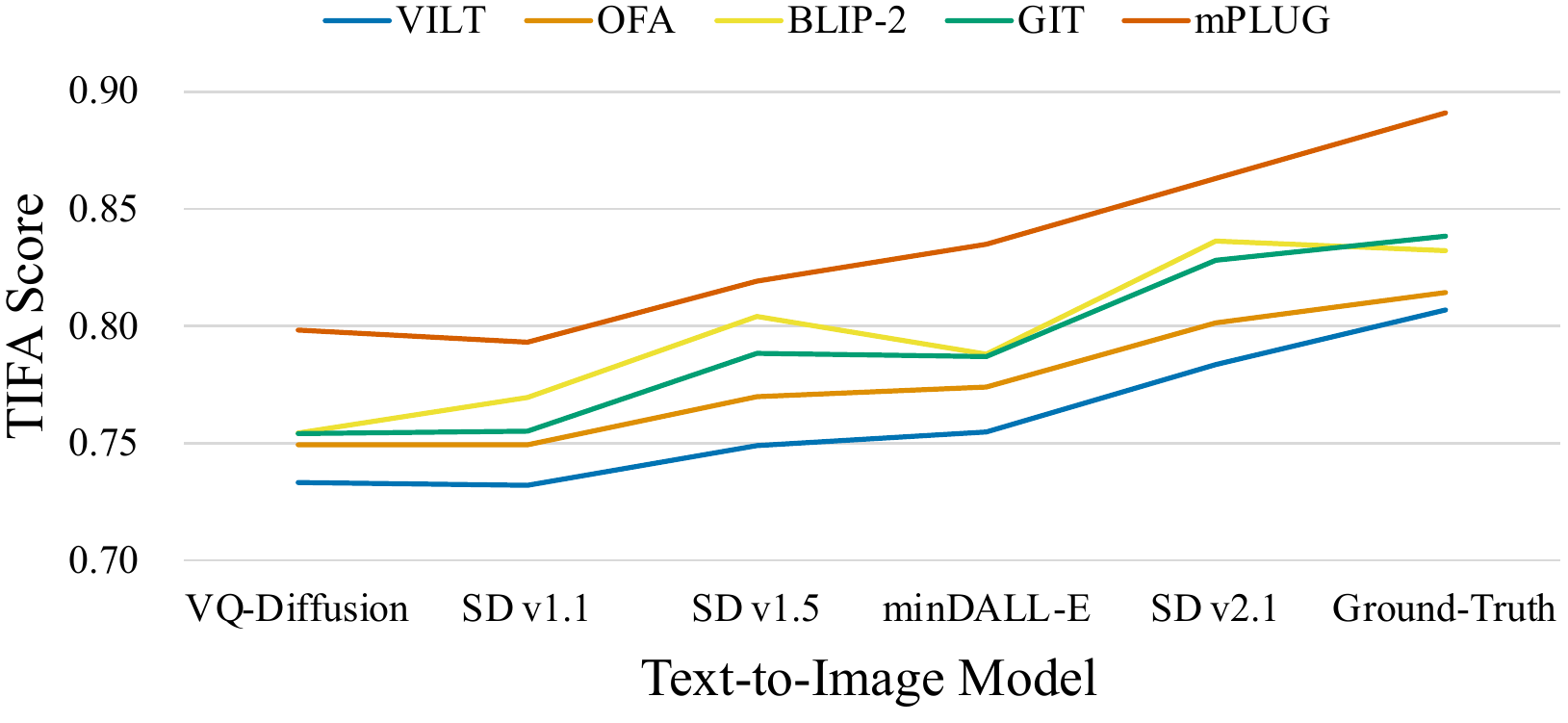}
  \caption{
Several text-to-image models' \NAME score on COCO captions, measured by different VQA models. We also include the accuracy of ground-truth images for reference. 
}
  \vspace{-4mm}
  \label{fig:vqa_models}
\end{figure}

\vspace{0.1in}\noindent\textbf{Sensitivity of \NAME to VQA models}
Figure~\ref{fig:vqa_models} shows several recent text-to-image models' \NAME scores on the COCO captions in \NAME v1.0, measured by different VQA models.
We also include the \NAME scores on the ground-truth COCO images for reference.
\NAME scores computed by different VQA models show a similar trend on these text-to-image models.
Also, the ground-truth images get the highest \NAME score.
We also computed Spearman's $\rho$ of \NAME scores given by different VQA models.
The pairwise correlation between all VQA models is greater than 0.6. 

\vspace{0.1in}\noindent\textbf{Humans performing VQA}
\label{sec:exp:humanVQA}
To conduct further analysis on the VQA models, we ask annotators to answer the multiple-choice visual questions in \NAME v1.0. These annotations help us evaluate the accuracy of each VQA model. For multiple-choice questions, we add the option ``None of the above" for human evaluation.
Annotation guidelines are in Appendix~\ref{appendix:annotation}.

We collect annotations of 1029 questions on 126 images. Each question is answered by two annotators. The inter-annotator agreement measured by Krippendorf's $\alpha$ is 0.88. A third annotator is involved if two annotators disagree, and the final answer is chosen by the majority vote.

\begin{table}[h]
\small
\centering
\caption{
Comparison of VQA models. The first row is the VQA accuracy, using the human VQA answers as reference. The second row is Spearman's correlation between \NAME scores calculated by each VQA model and the human VQA.
}

\begin{tabular}{lccccc}
\toprule[1.2pt]
&  VILT & OFA & GIT & BLIP-2 & mPLUG\\
\midrule
VQA Acc. & 76.1 & 77.1  & 79.1 & 81.0 & 84.5 \\
\NAME Corr. & 60.9 & 63.7 & 72.5 & 75.6 & 76.8 \\
\bottomrule[1.2pt]
\end{tabular}
\label{tab:vqa}

\end{table}

\vspace{0.1in}\noindent\textbf{Which VQA model should we use?}

Table~\ref{tab:vqa} reports the accuracy of each VQA model and the correlation between \NAME scores calculated by VQA model answers and human answers.  We observe that 
 higher model performance is directly related to the \NAME score's correlation with human judgments.
\textbf{mPLUG} has the highest accuracy.

Another important factor to consider is the runtime. We measure the inference speed of each VQA model on NVIDIA A40 GPU with batch size 1 over the Stable Diffusion v2.1 images ($768 \times 768$ pixels).
For one question, VILT takes 0.08s on average; OFA, GIT, and mPLUG all take about 0.25s; BLIP-2 takes 0.73s.
Based on the above results, we choose \textbf{mPLUG} as the default VQA model for \NAME v1.0 because it is the most accurate while being reasonably fast.

\vspace{0.1in}\noindent\textbf{Separation of Text-to-Image Errors and VQA Errors }
\label{sec:exp:vqa_errors}
Suppose an image gets a wrong answer given a visual question.
Then the image generation or the VQA model might have made an error.
Based on the human VQA results, we separate these two kinds of errors in Figure~\ref{fig:separate_error}.
If human VQA gives the wrong answer, then we suspect the generated image has an error. Otherwise, the image is correct but the VQA model is making an error.
Figure~\ref{fig:separate_error} shows that the majority of errors are made by the text-to-image models.
For mPLUG, less than 25\% errors are due to the VQA model.
This suggests that the \NAME framework is a viable evaluation method despite its inherent challenges.


\begin{figure}[ht]
\centering
  \includegraphics[width=0.45\textwidth]{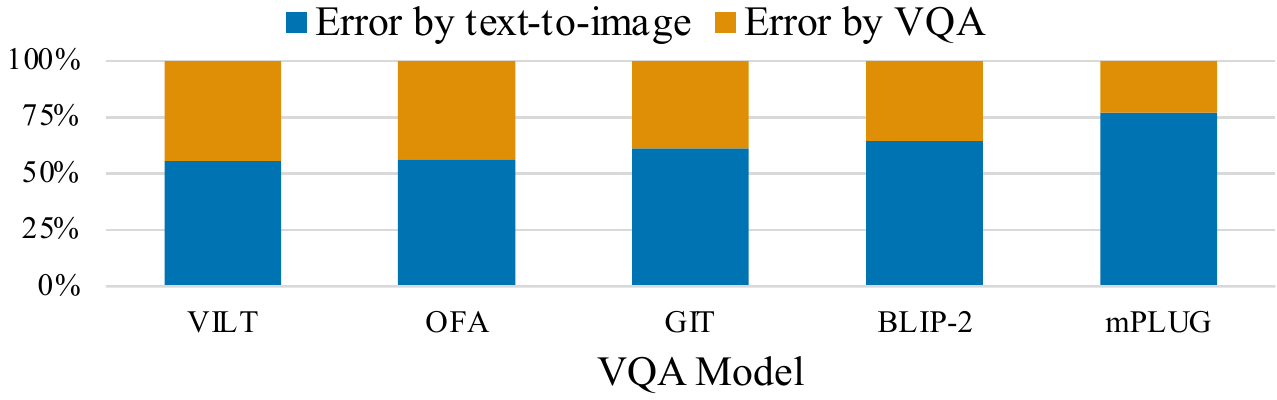}
  \caption{
Source of the error when VQA gets the wrong answer.
}
  \vspace{-4mm}
  \label{fig:separate_error}
\end{figure}

\section{Discussion}

\noindent\textbf{Why does \NAME work better than CLIPScore?}
Our experimental results and human evaluations in Section~\ref{sec:exp} show that \NAME is more accurate than prior metrics (CLIP~\cite{radford2021learning} and captioning-based approaches) for evaluating text-to-image faithfulness. We hypothesize that the major challenge of these prior metrics is that they summarize the image outputs and text inputs into a single representation (embedding/caption). In contrast, \NAME exploits the power of the language models to decompose the text input into fine-grained probes, which allows VQA to capture more nuanced aspects of the text input and the generated image.

\noindent\textbf{How do current VQA models perform on \NAME?}
One limitation is that \NAME requires VQA models to work reasonably well, which is true given the current models and \NAME v1.0, as shown in Section~\ref{sec:exp}. 
Nevertheless, the assumption might not hold for current models in domains like anime and abstract art.
TIFA is a modularized evaluation framework.
The VQA models used within the framework can be updated as stronger VQA models become available in the future. 
For instance, we plan to incorporate GPT-4 once its image API is made public since it is likely to improve TIFA. Another possible solution is to ensemble multiple image understanding models. For example, one may employ expert models on art concepts. We leave this for future work.

\vspace{1mm}
\noindent\textbf{Other limitations.}
Another limitation of \NAME is its runtime. Answering multiple visual questions is slower than one CLIP inference. In the scenario described in Section~\ref{sec:exp:comparisonVQA}, mPLUG takes 1.6s to evaluate one image (without batching). Also, our question generation pipeline needs one inference on a modern language model for each text input.
The run time is not a critical issue for benchmarking purposes, but may not be computationally feasible for the kind of large-scale data filtering done, for example, in LAION-5B~\cite{schuhmann2022laion}.
Nevertheless, we would like to point out that our evaluation is much faster than the image generation process of diffusion models.
Thus, we believe it is feasible to perform reranking and reinforcement learning with \NAME on diffusion models.

\section{Conclusions}
We present \NAME, a new automatic text-to-image faithfulness evaluation metric using VQA. Compared with prior metrics, \NAME is fine-grained, interpretable, and better aligned with human judgments. Based on this metric, we introduce the \NAME v1.0, a large-scale text-to-image benchmark containing 4K prompts and 25K questions. We conduct a comprehensive study of current text-to-image models using \NAME v1.0 and highlight the limitations of current generative models.
We quantitatively show that current image generation models still struggle in counting, spatial relations, and composing multiple objects.
Finally, we conduct extensive analysis and human evaluation, demonstrating that \NAME is robust to different VQA models.
We hope \NAME will help evaluate future work on image generation and become increasingly sophisticated as it is upgraded with new LM, QA, and VQA components.

\section{Acknowledgements}
We thank Zirong Ye, Cheng-Yu Hsieh, Oscar Michel, Enhao Zhang, Jiafei Duan, Weijia Shi, Jieyu Zhang, the TIAL group, the ARK group,  and the GRAIL lab at UW for their helpful feedback on this work. We would also thank all anonymous reviewers for providing insightful and constructive comments on our work.

{\small
\bibliographystyle{ieee_fullname}
\bibliography{references}
}

\clearpage
\appendix
\begin{appendices}

\section{Qualitative Examples}
\label{appendix:example}

Here we show three examples on how \NAME is computed. For each text input, we show the questions and answers generated by GPT-3~\cite{brown2020language} and filtered by UnifiedQA~\cite{Khashabi2020UnifiedQACF}. 
We also show the VQA model's answer to each vision question given each generated image.
The VQA model used is mPLUG~\cite{Li2022mPLUGEA}. The first text input comes from COCO~\cite{lin2014microsoft} and the second and third text inputs come from DrawBench~\cite{Saharia2022PhotorealisticTD}.

\begin{figure}[ht]
\centering
  \includegraphics[width=0.37\textwidth]{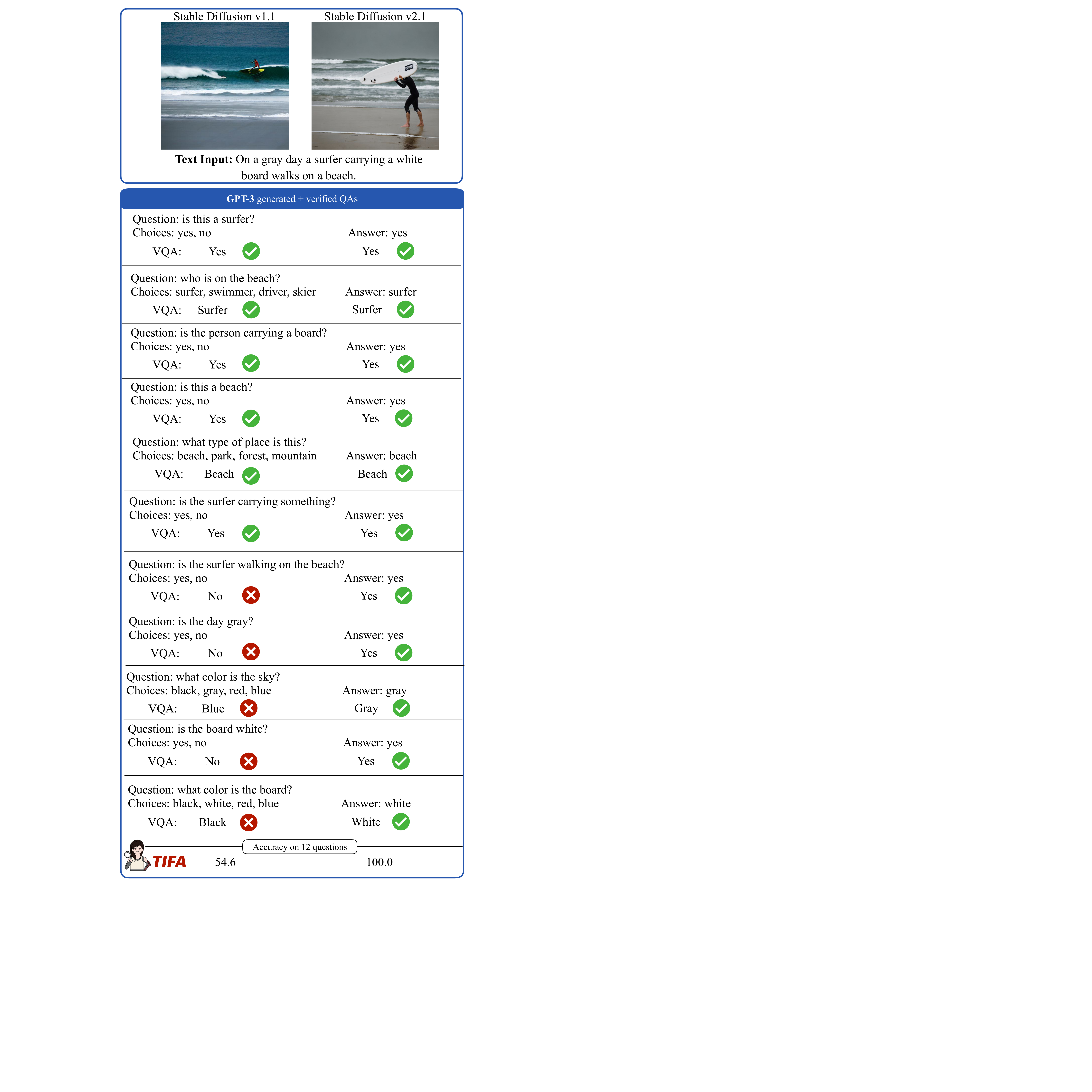}
  \label{fig:separate_error}
\end{figure}

\begin{figure}[ht]
\centering
  \includegraphics[width=0.37\textwidth]{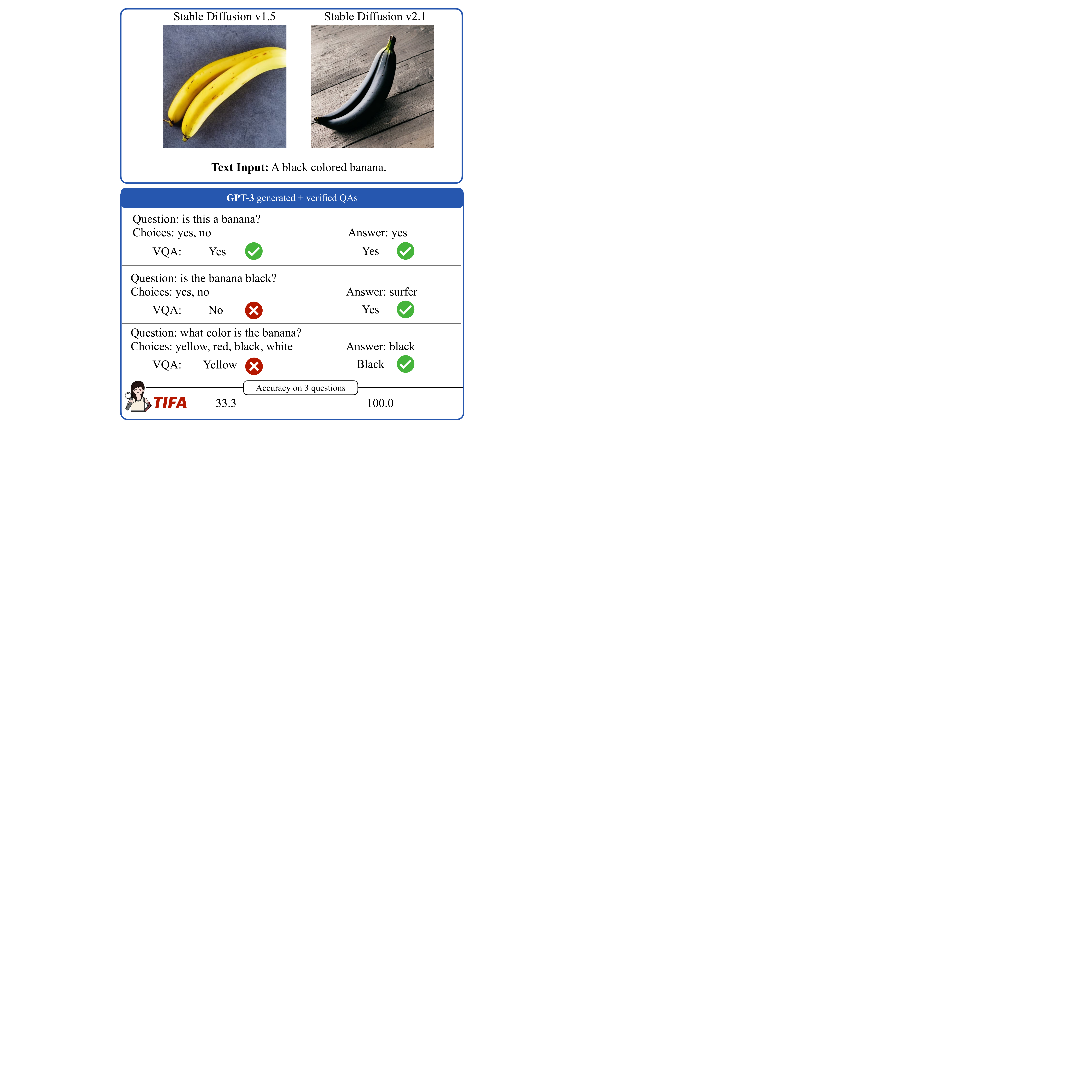}
  \label{fig:separate_error}
\end{figure}

\begin{figure}[ht]
\centering
  \includegraphics[width=0.37\textwidth]{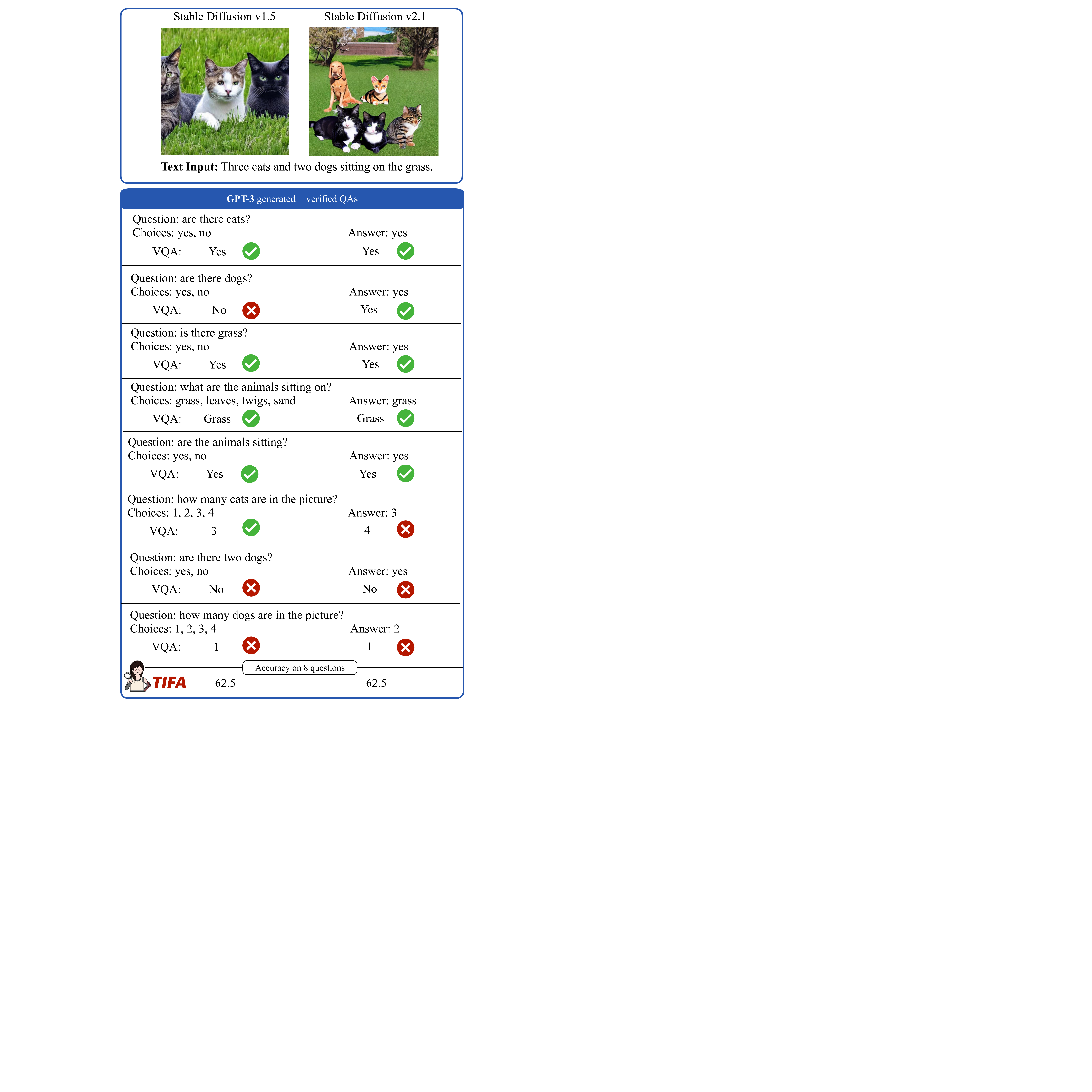}
  \label{fig:separate_error}
\end{figure}

\pagebreak

\section{Detailed Results}
\label{appendix:results}
\begin{table*}[h]
\small
\centering
\caption{
Detailed evaluation of each text-to-image model on \NAME v1.0.
}

\begin{tabular}{l|cccccccc}
\toprule[1.2pt]
&  \multicolumn{8}{c}{VQA Accuracy by Element Category}\\
& shape & other & counting & spatial & attribute & activity & food &object \\
\midrule
AttnGAN\cite{Xu2017AttnGANFT} & 42.0 & 47.8 & 41.9 & 70.8 & 53.6 & 64.3 & 48.1 & 56.3\\
X-LXMERT\cite{Cho2020XLXMERTPC} & 34.8 & 46.8 & 41.7 & 70.9 & 55.2 & 65.4 & 52.4 & 57.0\\
Stable Diffusion v1.1\cite{Rombach2021HighResolutionIS} & 66.7 & 68.7 & 66.0 & 69.4 & 74.6 & 73.8 & 79.7 & 75.1 \\
VQ-Diffusion\cite{Gu2021VectorQD} & 63.8 & 64.2 & 61.6 & 73.5 & 75.4 & 76.7 & 80.0 & 74.2\\
Stable Diffusion v1.5\cite{Rombach2021HighResolutionIS} & 65.2 & 72.1 & 66.6 & 72.9 & 78.0 & 76.9 & 81.4 & 78.4 \\
minDALL-E\cite{kakaobrain2021minDALL-E} & \textbf{69.6} & \textbf{74.6} & 69.0 & 74.7 & 77.0 & 79.5 & \textbf{82.8} & 79.9\\
Stable Diffusion v2.1\cite{Rombach2021HighResolutionIS} & 66.7 & 72.1 & \textbf{73.3} & \textbf{76.1} & \textbf{78.8} & \textbf{82.0} & 82.2 & \textbf{82.4} \\
\bottomrule[1.2pt]
\end{tabular}

\vspace{0.2in}

\begin{tabular}{l|cccc|cc|c}
\toprule[1.2pt]
&  \multicolumn{4}{c}{} & \multicolumn{2}{c}{\NAME by text source} & \textbf{Overall \NAME}\\
& location & color & animal/human & material & COCO & free-form & \\
\midrule
AttnGAN\cite{Xu2017AttnGANFT} & 60.4 & 56.5 & 58.6 & 61.7 & 67.5 & 47.4 & 58.1\\
X-LXMERT\cite{Cho2020XLXMERTPC} & 69.1 & 54.8 & 52.8 & 61.2 & 68.1 & 47.7 & 58.6\\
Stable Diffusion v1.1\cite{Rombach2021HighResolutionIS} & 78.4 & 75.7 & 78.2 & 80.4& 79.3 & 72.2 & 75.7\\
VQ-Diffusion\cite{Gu2021VectorQD} & 77.9 & \textbf{84.2} & 79.0 & 80.9 & 79.8 & 72.6 & 76.2 \\
Stable Diffusion v1.5\cite{Rombach2021HighResolutionIS} & 79.9 & 78.8 & 80.6 & 84.7 & 81.9 & 74.9 & 78.4 \\
minDALL-E\cite{kakaobrain2021minDALL-E} & 82.1 & 83.7 & 78.9 & 86.1 & 83.5 & 75.5 & 79.4 \\
Stable Diffusion v2.1\cite{Rombach2021HighResolutionIS} & \textbf{82.8} & 83.6 & \textbf{85.2} & \textbf{88.5} & \textbf{86.3} & \textbf{77.7} & \textbf{82.0}\\
\bottomrule[1.2pt]
\end{tabular}

\label{tab:detailed_scores}
\vspace{-4mm}
\end{table*}

Table~\ref{tab:detailed_scores} shows the detailed evaluation results of the text-to-image models we use on the \NAME v1.0 benchmark.
We show the VQA accuracy of each question category, \NAME score on each text source, and the overall \NAME score. 
We can see that Stable Diffusion v2.1~\cite{Rombach2021HighResolutionIS} gets the highest overall score and also scores the highest in most categories. Nonetheless, the CLIP~\cite{radford2021learning} and VQGAN~\cite{Esser2020TamingTF} based minDALL-E~\cite{kakaobrain2021minDALL-E} gets the highest accuracy on ``shape", ``other", ``food", and VQ-Diffusion~\cite{Gu2021VectorQD} gets the highest accuracy on ``color".

\section{Annotation Details}
\label{appendix:annotation}

\subsection{Likert Scale on Text-to-Image Faithfulness}
\paragraph{Guidelines}
The annotation guideline is as follows:
\begin{itemize}
    \item On a scale of 1-5, score "does the image match the prompt?".
    \item The ranking of each image given the same text input is important. If you believe the current scoring criteria cannot reflect your ranking preference, pick scores that are consistent with your ranking. Ties are allowed.
    \item To evaluate the generated image, there are two aspects: image quality and text-image match. Here we only care about text-image match, which is referred to as “faithfulness”.
    \item There are several kinds of elements in the text: object, attribute, relation, and context. Measure the consistency by counting how many elements are missed/misrepresented in the generated image.
    \item For some elements, e.g. ``train conductor's hat", if you can see there is a hat but not a train conductor's hat, consider half of the element is missed/misrepresented in the generated image.
    \item Objects are the most important elements. If an object is missing, then consider all related attributes, activity, and attributes missing.
    \item When you cannot tell what the object/attribute/activity/context is, consider the element missing. (e.g., can't tell if an object is a microwave)
\end{itemize}

Given the above guideline, suppose the text input contains $n$ elements, and $x$ elements are missed or misrepresented. 
$n$ and $x$ are all counted by the annotators.
The reference scoring guideline is as follows:
\begin{itemize}
    \item 5: The image perfectly matches the prompt.
    \item 4:  $x \le 2$ and $x \le n/3$. A few elements are missed/misrepresented.
    \item 3: $\min \{2, n/3\} < x \le n/2$ elements are missed/misrepresented.
    \item 2: $x > n/2$. More than half of the elements are missed/misrepresented.
    \item 1: None of the major objects are correctly presented in the image.
\end{itemize}

\paragraph{Details}
We collect 1600 annotations on 800 generated images from 160 text inputs. Each image is scored by 2 annotators, and we collect the scores from 20 graduate students. We average the scores as the final faithfulness score of the image. The inter-annotator agreement measured by Krippendorf's $\alpha$ is 0.67, indicating ``substantial" agreement.
The images are generated by the five most recent text-to-image models in our study, including VQ-Diffusion~\cite{Gu2021VectorQD}, minDALL-E~\cite{kakaobrain2021minDALL-E}, and Stable Diffusion~\cite{Saharia2022PhotorealisticTD} v1.1, v1.5, and v2.1. 
For each text input, we present the five images together, making it easier for the annotators to give faithfulness scores that reflect their ranking preference.
We will release the annotation scores on publication.

\subsection{Human VQA}

\paragraph{Guidelines}
Given an image, a question, and a set of choices, choose the correct choice according to the image content. There are two types of questions. One has two options: "(A) yes (B) no". Another type of question has four choices. We also add the fifth option ``None of the above". If you believe none of the four choices is correct, choose the fifth one. 
Some images are of low quality. Just select the choice according to your instinct.
For ambiguous cases, for example, the question is ``is there a man?", and the image contains a human but it is unclear whether the human is a man, answer ``no".

\paragraph{Details}
We collect annotations of 1029 questions on 126 generated images. The images are from images used in the Likert Scale annotation. Each question is answered by two annotators, and we have the same 20 graduate students as the annotators. The inter-annotator agreement measured by Krippendorf's $\alpha$ is 0.88. A third annotator is involved if the two annotators disagree. And the final answer is given by the majority vote. We will release the annotated VQA answers.

\section{Common Q \& A}
\label{appendix:qa}

\paragraph{Any possible extension to \NAME?}
As discussed in \S\ref{sec:intro}, one extension of our work will be customized versions of the \NAME benchmark focusing on one aspect of image generation. For example, we can make a \NAME benchmark that only contains questions about ``counting"; Or a benchmark consists of text inputs synthesized to test text-to-image models' ability in composing multiple objects. Another possible extension is to use \NAME on other generation tasks, e.g., text-to-3D and text-to-video.

\paragraph{The OpenAI APIs are too expensive. Can we generate questions by local models?}
Yes. Our approach works on any language model. Please refer to \S\ref{sec:question_generation_llama} on question generation with our fine-tuned LLaMA 2 checkpoint in.  Also, we would like to emphasize that all questions in \NAME v1.0 benchmark are pre-generated by GPT-3, and there is no need to re-generate those questions for evaluation.

\begin{figure}[h]
\centering
  \includegraphics[width=0.4\textwidth]{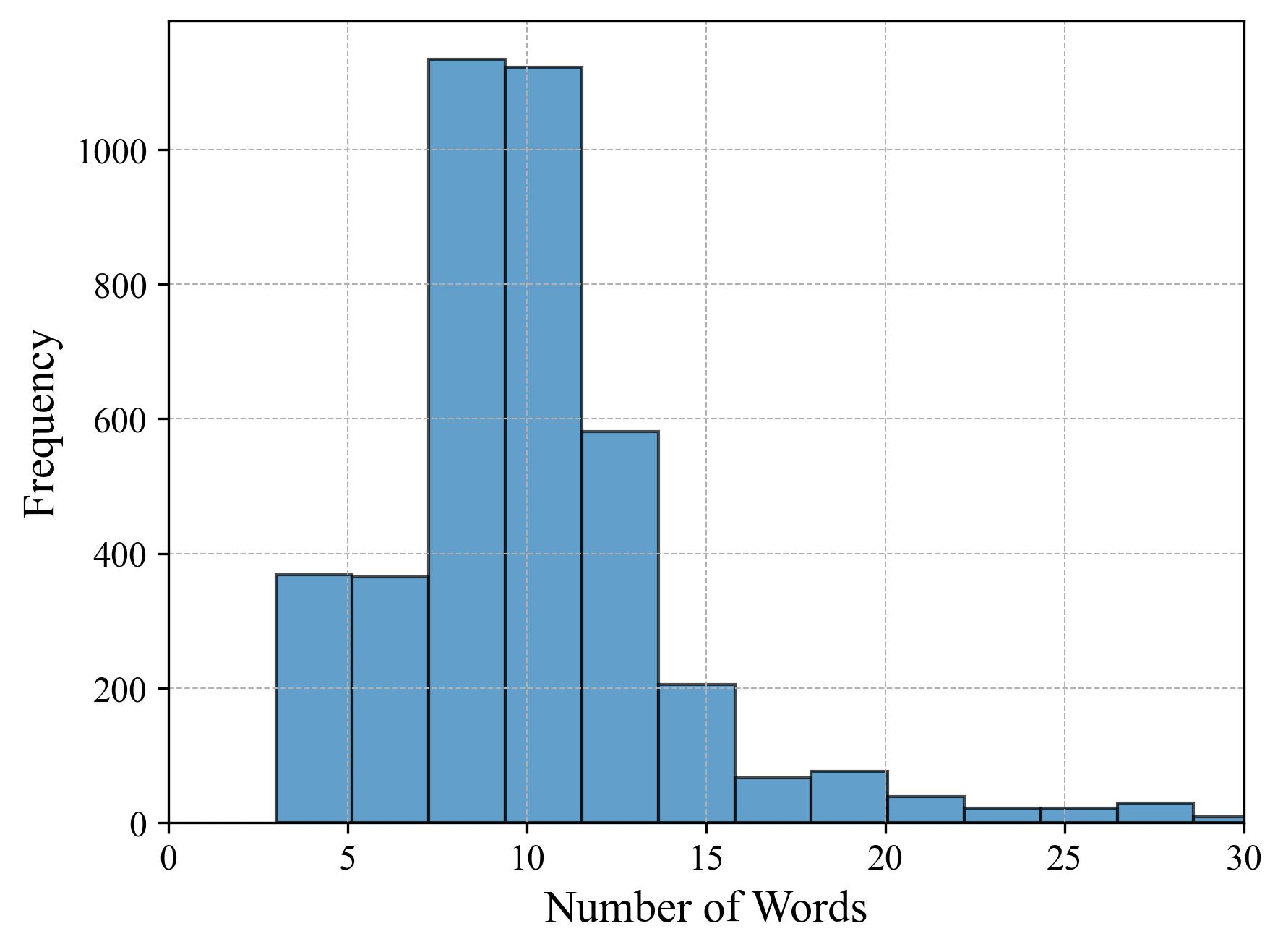}
  \caption{
Distribution of the lengths of \NAME v1.0 text inputs.
}
  \vspace{-4mm}
  \label{fig:tifa_len}
  
\end{figure}

\paragraph{More details on \NAME v1.0 text inputs?}
The distribution of the number of words in the text inputs is shown in ~\ref{fig:tifa_len}. Most text inputs have around 10 words. We also conduct bias analysis on \NAME v1.0 text inputs. Among the 4K text inputs, regarding gender expression, 400 are perceived as ``male" and 239 are perceived as ``female". We find that the bias in gender distribution comes from captions sampled from COCO dataset~\cite{lin2014microsoft}.

\section{Text-to-Image Model Details}
\label{appendix:t2i}

\paragraph{AttnGAN} 
AttnGan~\cite{Xu2017AttnGANFT} is a text-to-image model that is introduced in 2017. It is based on an attention mechanism that allows the model to focus on different parts of the input text when generating an image. AttnGAN has been shown to generate high-quality images for a variety of datasets and has been widely used in a number of applications. However, the attention mechanism can be computationally expensive, and the model can be difficult to train. 

\paragraph{X-LXMERT} 
X-LXMERT~\cite{Cho2020XLXMERTPC} is an enhanced version of LXMERT~\cite{Tan2019LXMERTLC}. It is introduced in 2020 and incorporates several training refinements. These refinements involve discretizing visual representations, utilizing uniform masking with a wide range of masking ratios, and aligning appropriate pre-training datasets to respective objectives.

\paragraph{minDALL-E}
MinDALL-E~\cite{kakaobrain2021minDALL-E} is a fast, minimal port of Boris Dayma's DALL·E Mini (with mega weights).
DALL·E Mini is an attempt to reproduce OpenAI's DALL-E~\cite{Ramesh2021ZeroShotTG} with a smaller architecture. DALL-E can generate high-quality new images from any text prompt. The checkpoint we use is DALL·E Mega, the latest version of DALL·E Mini.

\paragraph{VQ-Diffusion}
VQ-Diffusion~\cite{Gu2021VectorQD} is a generative model that combines vector quantization (VQ) and diffusion-based models for image synthesis. VQ-Diffusion builds upon the framework of diffusion-based generative models, which involves simulating a stochastic process that gradually transforms a simple noise distribution into the target data distribution. In VQ-Diffusion, the image data is first quantized into discrete codes using a VQ algorithm, which maps each image patch to the nearest code in a codebook. This allows the model to represent complex data distributions with a compact set of discrete codes, rather than continuous probability densities.

\paragraph{Stable Diffusion}
Stable Diffusion is a pre-trained diffusion model for text-to-image generation. It is based on Latent Diffusion model (LDM)~\cite{Rombach2021HighResolutionIS}. LDM is designed to learn the underlying structure of a dataset by mapping it to a lower-dimensional latent space. This latent space represents the data in which the relationships between different data points are more easily understood and analyzed, and reduces the amount of computational resources needed for training diffusion models. Specifically, we use three versions of Stable Diffusion, v1.1, v1.5, and v2.1. Each version is trained with a different number of steps and amount of data.

\section{Prompt}
\label{appendix:prompt}

For demonstration purposes, we show part of the prompt for question generation with GPT-3 in-context learning. The whole prompt will be released with our codes. The prompt contains instructions and several in-context examples. The examples cover all element categories.

\begin{tiny}
\begin{lstlisting}
Given an image description, generate multiple-choice questions that verify if the image description is correct.

First extract elements from the image description. Then classify each element into a category (object, human, animal, food, activity, attribute, counting, color, material, spatial, location, shape, other). Finally, generate questions for each element.

Description: A man posing for a selfie in a jacket and bow tie.
Entities: man, selfie, jacket, bow tie
Activities: posing
Colors:
Counting:
Other attributes:
Questions and answers are below:
About man (human):
Q: is this a man?
Choices: yes, no
A: yes
Q: who is posing for a selfie?
Choices: man, woman, boy, girl
A: man
About selfie (activity):
Q: is the man taking a selfie?
Choices: yes, no
A: yes
Q: what type of photo is the person taking?
Choices: selfie, landscape, sports, portrait
A: selfie
About jacket (object):
Q: is the man wearing a jacket?
Choices: yes, no
A: yes
Q: what is the man wearing?
Choices:jacket, t-shirt, tuxedo, sweater
A: jacket
About bow tie (object):
Q: is the man wearing a bow tie?
Choices: yes, no
A: yes
Q: is the man wearing a bow tie or a neck tie?
Choices: bow tie, neck tie, cravat, bolo tie
A: bow tie
About posing (activity):
Q: is the man posing for the selfie?
Choices: yes, no
A: yes
Q: what is the man doing besides taking the selfie?
Choices: posing, waving, nothing, shaking
A: posing

Description: A horse and several cows feed on hay.
Entities: horse, cows, hay
Activities: feed on
Colors:
Counting: several
Other attributes:
Questions and answers are below:
About horse (animal):
Q: is there a horse?
Choices: yes, no
A: yes
About cows (animal):
Q: are there cows?
Choices: yes, no
A: yes
About hay (object):
Q: is there hay?
Choices: yes, no
A: yes
Q: what is the horse and cows feeding on?
Choices: hay, grass, leaves, twigs
A: hay
About feed on (activity):
Q: are the horse and cows feeding on hay?
Choices: yes, no
A: yes
About several (counting):
Q: are there several cows?
Choices: yes, no
A: yes

Description: A red colored dog.
Entities: dog
Activities:
Colors: red
Counting:
Other attributes:
Questions and answers are below:
About dog (animal):
Q: is this a dog?
Choices: yes, no
A: yes
Q: what animal is in the picture?
Choices: dog, cat, bird, fish
A: dog
About red (color):
Q: is the dog red?
Choices: yes, no
A: yes
Q: what color is the dog?
Choices: red, black, white, yellow
A: red

Description: Here are motorcyclists parked outside a Polish gathering spot for women
Entities: motorcyclists, gathering spot, women
Activities: parked
Colors:
Counting:
Other attributes: outside, polish
Questions and answers are below:
About motorcyclists (human):
Q: are there motorcyclists?
Choices: yes, no
A: yes
About gathering spot (location):
Q: is this a gathering spot?
Choices: yes, no
A: yes
About women (human):
Q: are there women?
Choices: yes, no
A: yes
Q: who are in the gathering spot?
Choices: women, men, boys, girls
A: women
About parked (activity):
Q: are the motorcyclists parked?
Choices: yes, no
A: yes
About outside (spatial):
Q: have the motorcyclists parked outside the gathering spot?
Choices: yes, no
A: yes
Q: are the motorcyclists outside or inside of the gathering spot?
Choices: outside, inside, on the roof, in the basement
A: outside
About Polish (other):
Q: is this a Polish gathering spot?
Choices: yes, no
A: yes
Q: is this a Polish or a Chinese gathering spot?
Choices: Polish, American, Chinese, Japanese
A: Polish
\end{lstlisting}
\end{tiny}

\end{appendices}

\end{document}